\documentclass[]{article}
\usepackage{proceed2e}

\usepackage{url,mathrsfs}
\usepackage{amsmath,wrapfig,color,bigfoot}
\usepackage{amsfonts}
\usepackage{graphicx}
\usepackage{subfigure}
\newtheorem{theorem}{Theorem}

\title{\vspace{-0.1in}CoRE Kernels\vspace{-0.1in}}

\author{ {\bf Ping Li} \\
Department of Statistics and Biostatistics\\
Department of Computer Science \\
Rutgers University\\
Piscataway, NJ 08854, USA \\
\texttt{pingli@stat.rutgers.edu}
}

\begin{document}

\maketitle
\begin{abstract}\vspace{-0.2in}

The\footnote{The idea of combining permutation with  projection was developed in the proposal NSF-III1360971, which was funded after several rounds of submissions.}
term ``CoRE kernel'' stands for {\em correlation-resemblance kernel}. In many  applications (e.g., vision), the data are often high-dimensional, sparse, and non-binary. We propose two types of (nonlinear) CoRE kernels for non-binary sparse data and demonstrate the effectiveness of the new kernels through a classification experiment. CoRE kernels are  simple with no tuning parameters. However, training nonlinear kernel SVM can be (very) costly in  time and memory and may not be suitable for truly large-scale industrial applications (e.g. search). In order to make the proposed CoRE kernels more practical, we develop basic probabilistic hashing algorithms which  transform nonlinear kernels into linear kernels.

\end{abstract}

\vspace{-0.25in}
\section{Introduction}
\vspace{-0.1in}

The use of high-dimensional data has become very popular these days, especially in search, natural language processing (NLP), and computer vision.  For example, Winner of 2009 PASCAL image classification challenge  used {4 million} (non-binary) features~\cite{Proc:Wang_CVPR10}. \cite{Report:Sibyl,GoogleBlog,Proc:Weinberger_ICML2009} discussed datasets with billions or even trillions of features.

For text data, the use of extremely high-dimensional representations (e.g., $n$-grams) is the standard practice, for example, \cite{Report:Sibyl,GoogleBlog,Proc:Weinberger_ICML2009}. Binary representations could be sufficient if the order of $n$-grams is high enough. On the other hand, in  current practice of computer vision, it is still more common to use non-binary feature representations, for example, {\em local coordinate coding (LCC)}~\cite{Proc:LCC_NIPS09,Proc:Wang_CVPR10}. It is often the case that in practice high-dimensional non-binary visual features might be appropriately sparsified without hurting the performance. However simply binarizing the features will often incur loss of accuracies, sometimes significantly so.  See Table~\ref{tab_linearSVM} for an illustration.

Our contribution in this paper is the proposal of two types of (nonlinear) ``CoRE'' kernels, where ``CoRE'' stands for ``correlation-resemblance'', for non-binary sparse data. Interestingly, using CoRE kernels leads to improvement in classification accuracies (in some cases significantly so)  on a variety of datasets.

For practical large-scale applications, naive implementations of nonlinear kernels may be too costly (time and/or memory), while linear  learning methods (e.g., linear SVM or logistic regression) are extremely popular in  industry.  The proposed CoRE kernels would be facing the same challenge.  To address this critical issue, we also develop efficient hashing algorithms which approximate the  CoRE kernels by linear kernels. These new hashing algorithms allow us to take advantage of  highly efficient (batch or stochastic) linear algorithms, e.g.,~\cite{Proc:Joachims_KDD06,Proc:Shalev-Shwartz_ICML07,URL:Bottou_SGD,Article:Fan_JMLR08}.

In the rest of this section, we first review the definitions of correlation and resemblance, then we provide an experimental study to illustrate the loss of classification accuracies when  sparse data are binarized.

\vspace{-0.1in}
\subsection{Correlation}
\vspace{-0.1in}

We assume a data matrix of size $n\times D$, i.e., $n$ observations in $D$ dimensions. Consider, without loss of generality, two data vectors $u, v\in\mathbb{R}^D$.   The correlation is simply the normalized inner product:
\begin{align}\label{eqn_rho}
&\rho = \rho(u,v) = \frac{\sum_{i=1}^D u_i v_i}{\sqrt{\sum_{i=1}^D u_i^2\sum_{i=1}^D v_i^2}} = \frac{A}{\sqrt{m_1 m_2}},\\\notag
&\text{where } \ A = \sum_{i=1}^D u_iv_i, \ \
m_1 = \sum_{i=1}^D u_i^2,\ \ m_2 = \sum_{i=1}^D v_i^2
\end{align}
It is well-known that $\rho(u,v)$  constitutes a positive definite and linear kernel, which is one of the reasons why correlation is very popular in practice.

\subsection{Resemblance}
For binary data, the resemblance is commonly used:
\begin{align}\label{eqn_R}
&R = R(u,v) = \frac{a}{f_1+f_2-a},  \\\notag
\text{where }\ &f_1 = \sum_{i=1}^D 1\{u_i \neq 0\},\hspace{0.2in}
f_2 = \sum_{i=1}^D 1\{v_i \neq 0\},\\\notag
 &a = \sum_{i=1}^D 1\{u_i \neq 0\} 1\{v_i\neq 0\}
\end{align}
It was shown in~\cite{Proc:HashLearning_NIPS11} that the resemblance defines a type of positive definite kernel. In this study, we will combine correlation and resemblance to define two new types of nonlinear kernels.

\vspace{-0.1in}
\subsection{Linear SVM Experiment}
\vspace{-0.05in}

Table~\ref{tab_linearSVM} lists the datasets, which are  non-binary and sparse. The table also presents the test classification accuracies using linear SVM on both the original (non-binary) data and  binarized data. The results in the table illustrate the noticeable drop of accuracies by using only binary data.

\begin{table}[h!]
\caption{Classification accuracies (in \%) using linear SVM (LIBLINEAR~\cite{Article:Fan_JMLR08}) on  sparse non-binary datasets. As we always normalize the data (to unit norm),  the correlation kernel $\rho$ is naturally used here. We experiment with the $l_2$-regularized linear SVM and report the best test accuracies from a wide range of ``$C$'' values (where $C$ is the parameter in linear SVM). Using binarized data (i.e., the last column), the test accuracies drop quite noticeably in most cases. \\
Available at the UCI repository, \textbf{Youtube}  is a multi-view dataset, and we choose the largest set of features (audio) for our experiment. \textbf{M-Basic}, \textbf{M-Rotate}, and \textbf{MNIST10k} were used in~\cite{Proc:ABC_UAI10} for testing {\em abc-logitboost} and {\em abc-mart}~\cite{Proc:ABC_ICML09} (and comparisons with deep learning~\cite{Proc:Larochelle_ICML07}). For \textbf{RCV1}, we use a subset of the original testing examples (to facilitate efficient kernel computation  later  needed in the paper).}
\begin{center}{
{\begin{tabular}{l r r r r r}
\hline \hline
Dataset        &\#Train    &\#Test &Linear &Lin. Bin.\\
\hline
M-Basic      &12,000 &50,000   &90.0\%   &88.9\% \\
MNIST10k     &10,000  &60,000  &90.0\%   &88.8\% \\
M-Rotate    &12,000 &50,000  &48.0\%   &44.4\%\\
RCV1       &20,242 &60,000 &96.3\%  &95.6\%\\
USPS        &7,291 &2,007 &91.8\%   &87.4\%\\
Youtube &11,930 &97,934 &47.6\% &46.5\%
\\\hline\hline
\end{tabular}}
}
\end{center}\label{tab_linearSVM}

\end{table}

\begin{figure}[h!]
\begin{center}
\mbox{\includegraphics[width=1.7in]{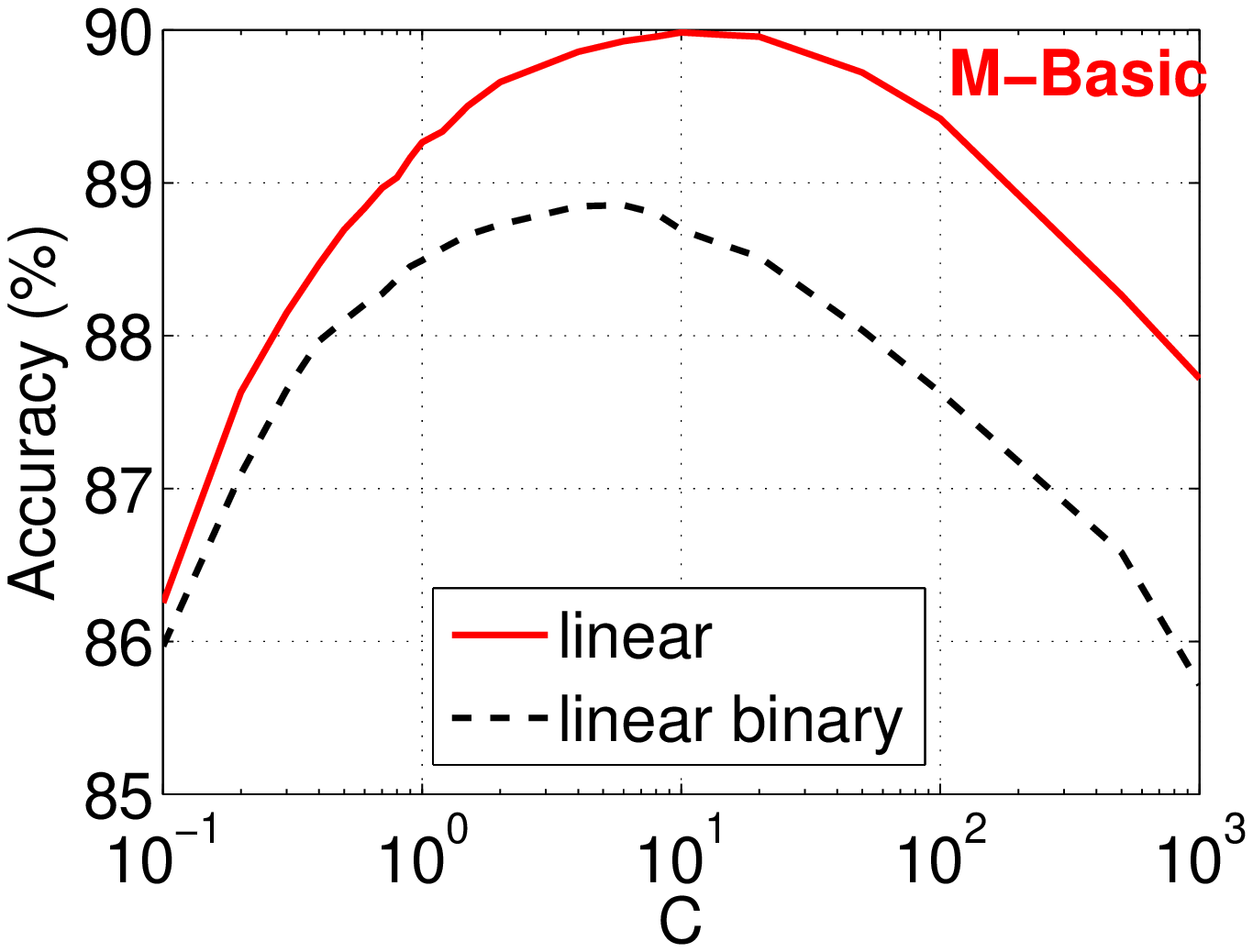}\hspace{-0.12in}
\includegraphics[width=1.7in]{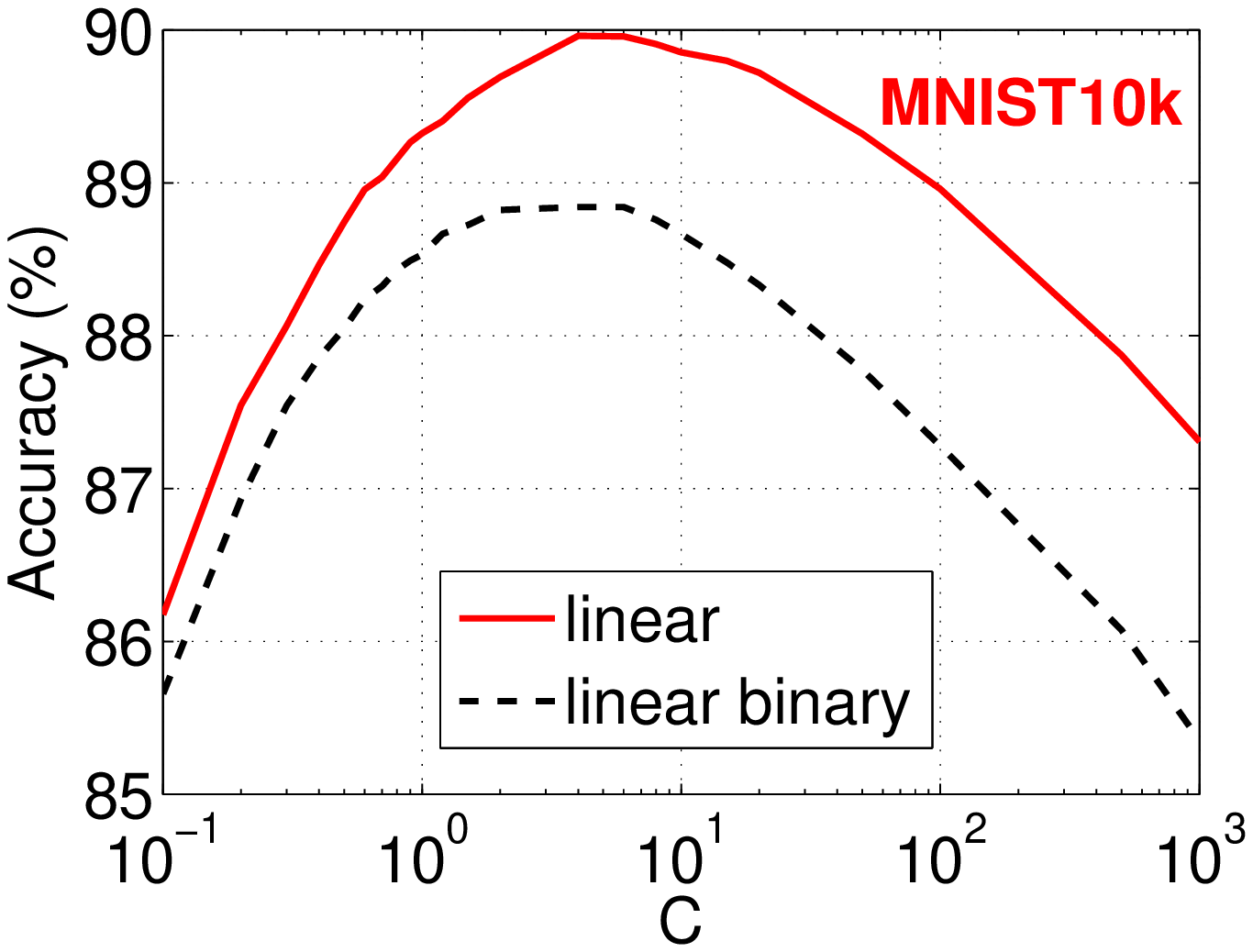}
}

\mbox{\includegraphics[width=1.7in]{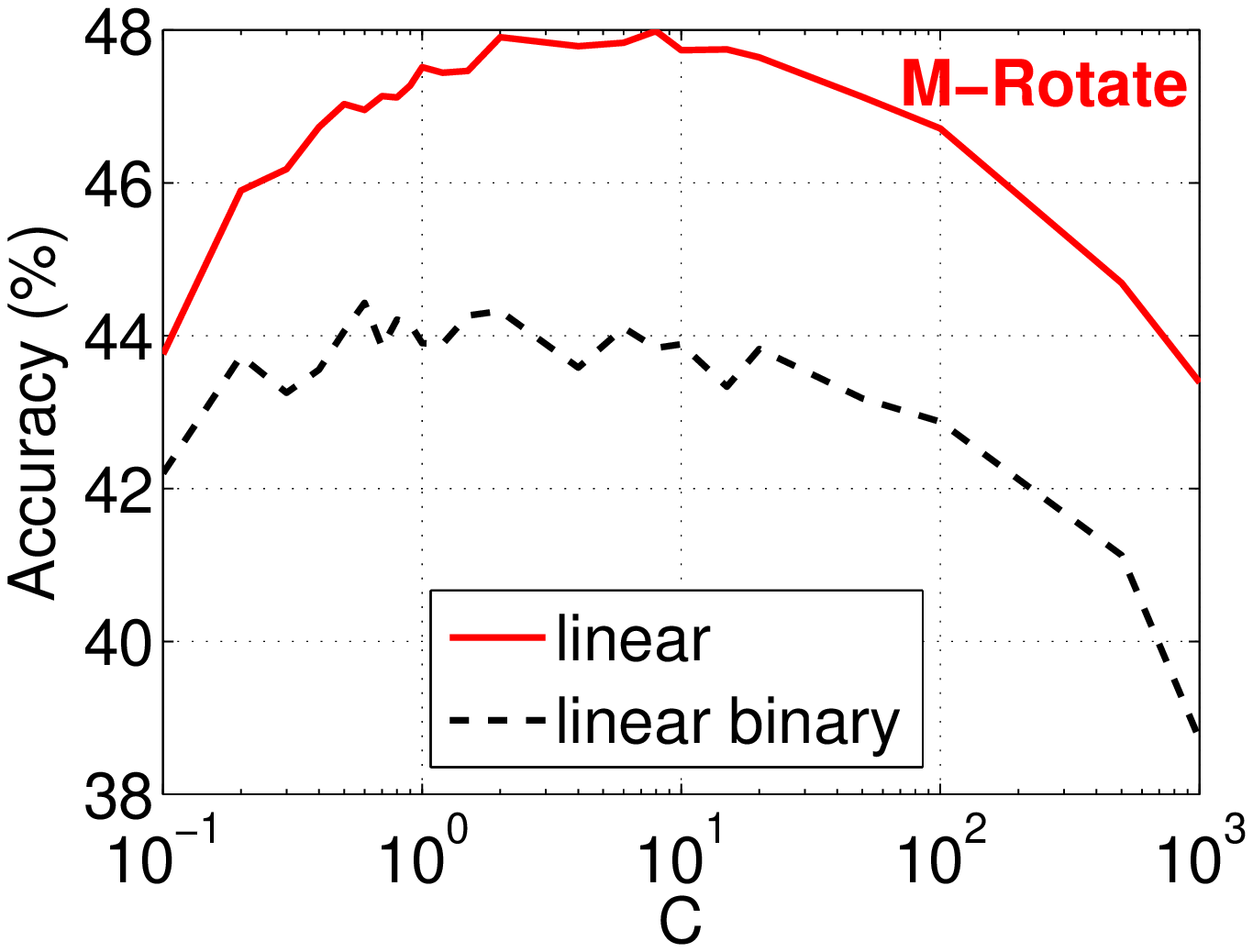}\hspace{-0.12in}
\includegraphics[width=1.7in]{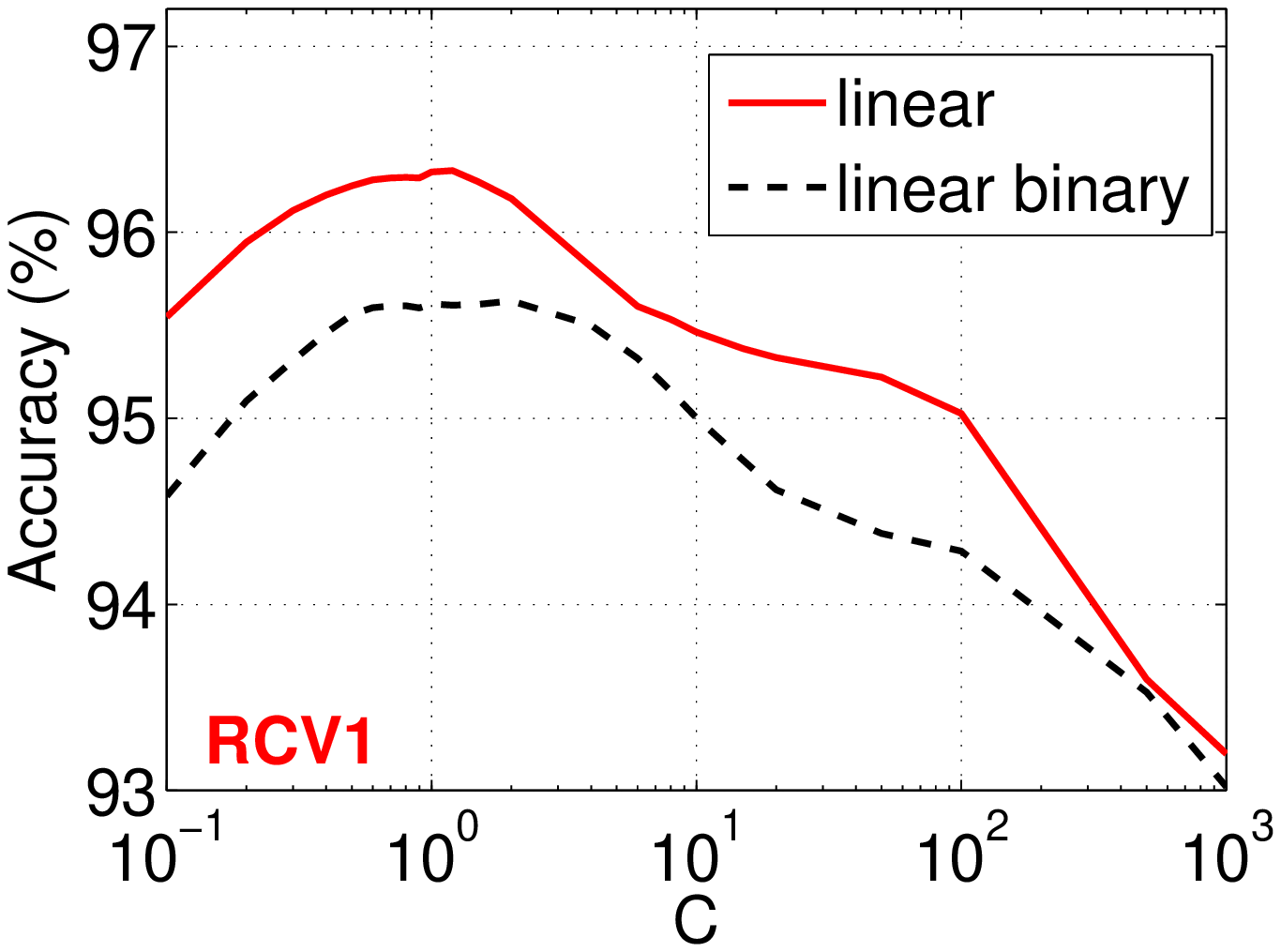}
}

\mbox{\includegraphics[width=1.7in]{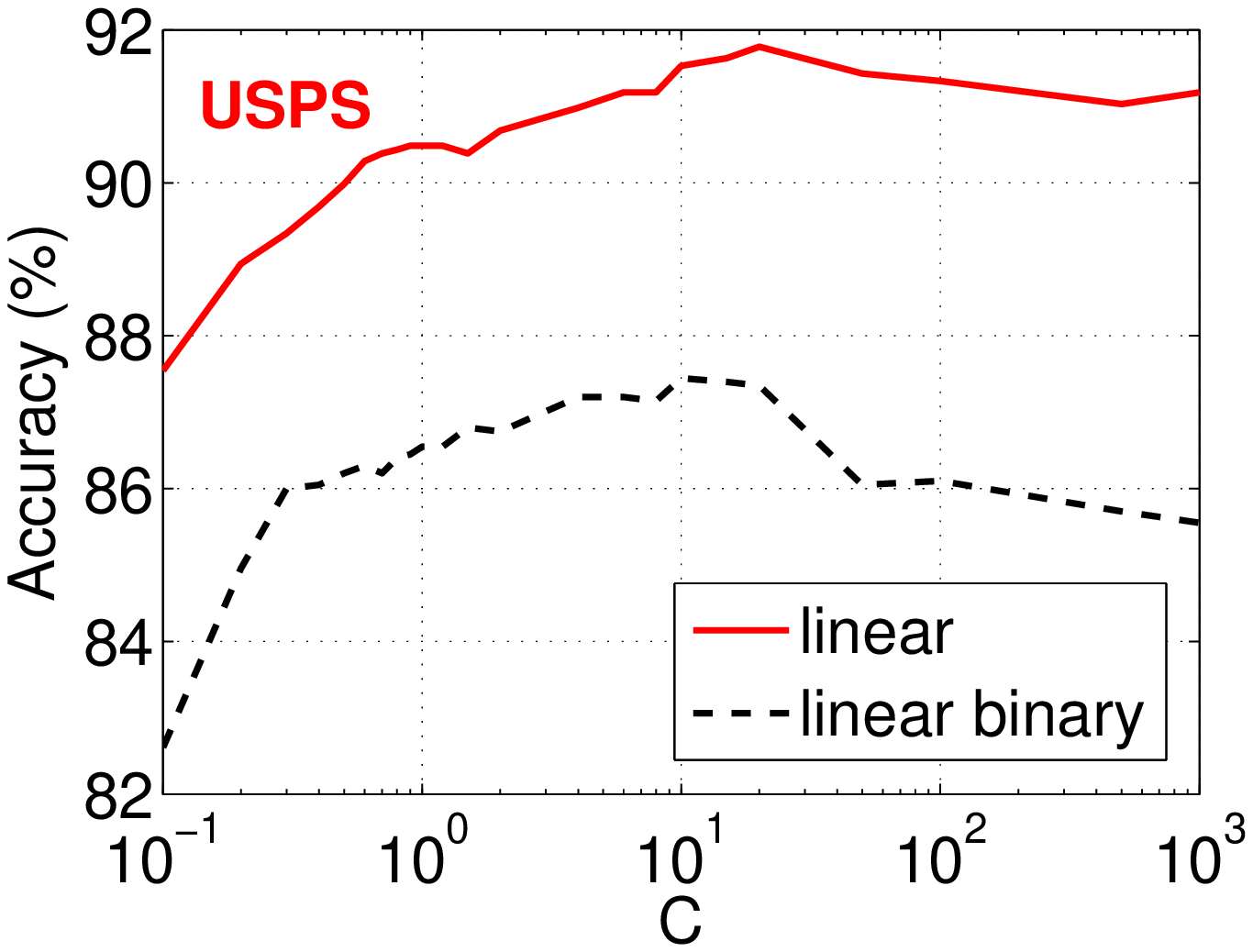}\hspace{-0.12in}
\includegraphics[width=1.7in]{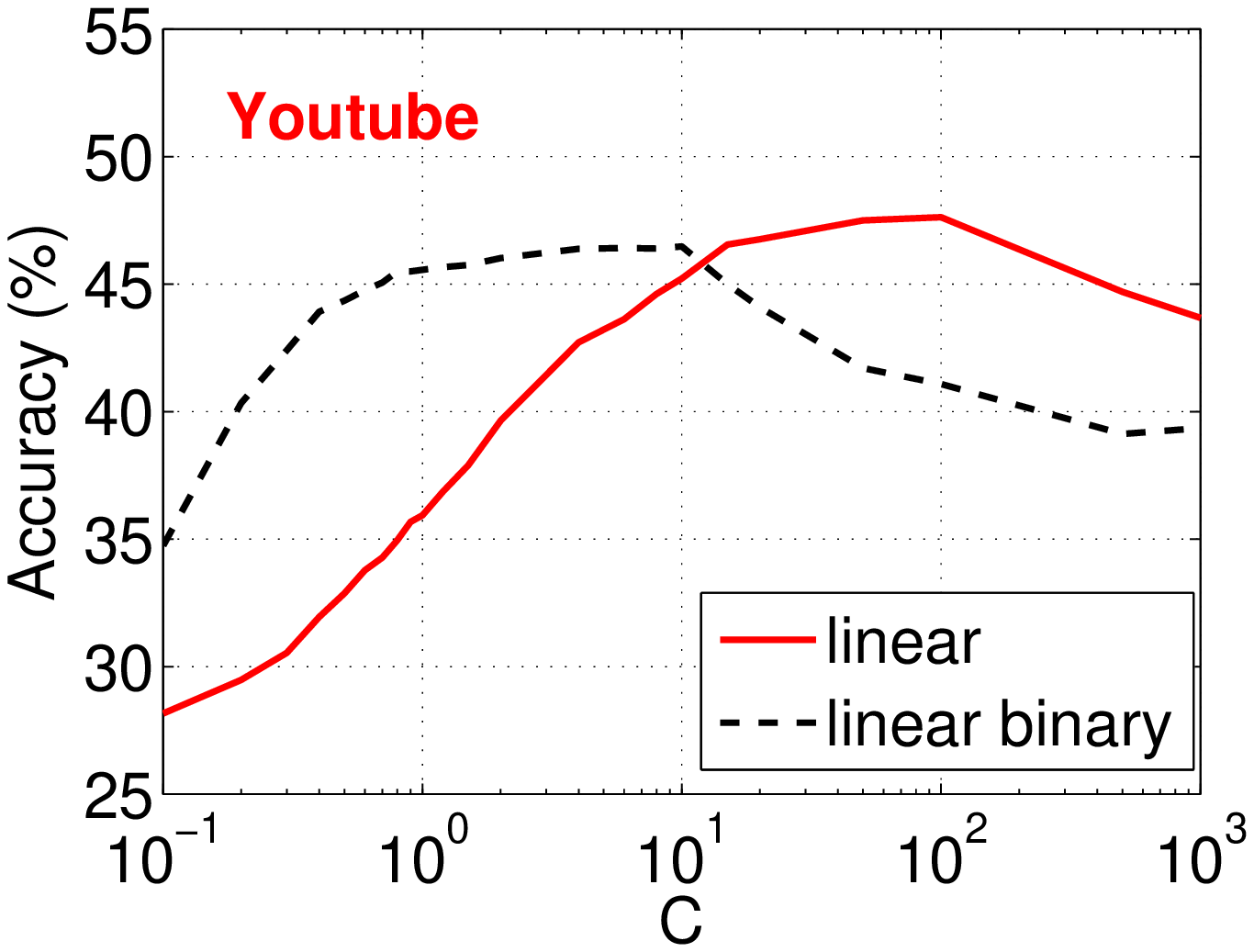}
}

\end{center}
\vspace{-0.2in}
\caption{Test classification accuracies for both non-binary (solid) and binarized (dashed) data, using $l_2$-regularized linear SVM with a regularization parameter $C$. We present results for a wide range of $C$ values. The best (highest) values are  summarized in Table~\ref{tab_linearSVM}.}\label{fig_linearSVM}\vspace{-0.1in}
\end{figure}

Figure~\ref{fig_linearSVM} provides more detailed results for a wide range of $C$ values, where $C$ is the usual $l_2$-regularization parameter in SVM.

While linear SVM is extremely popular in industrial practice, it is often not as accurate. Our proposed CoRE kernels will be able to produce noticeably more accurate results than linear SVM.

\vspace{-0.1in}
\section{CoRE Kernels}
\vspace{-0.05in}

We propose two types of CoRE kernels, which combine resemblance with correlation, for sparse non-binary data. Both kernels are positive definite. We will demonstrate the effectiveness of the two CoRE kernels using the same datasets in Table~\ref{tab_linearSVM} and Figure~\ref{fig_linearSVM}.\footnote{In order to use LIBSVM precomputed kernel functionality we need to materialize the full kernel matrix, which is extremely expensive to store. Also, it looks the original LIBSVM code limits the size of  kernel matrix. To ensure reproducibility, this paper did not use larger datasets.  }

\vspace{-0.1in}
\subsection{CoRE Kernel, Type 1}
\vspace{-0.05in}

The first type of CoRE kernel is basically the product of correlation $\rho$ and the resemblance $R$.
\begin{align}
K_{C,1} = K_{C,1}(u,v) =  \rho R
\end{align}
Later  we will express $K_{C,1}$ as an (expectation of) inner product, i.e., $K_{C,1}$ is obviously positive definite.

If  the data are fully dense, then $R=1$ and $K_{C,1} = \rho$. On the other hand, if the data are binary, then $\rho = \frac{a}{\sqrt{f_1f_2}}$ and $K_{C,1} = \frac{a}{\sqrt{f_1f_2}}\frac{a}{f_1+f_2-a}$. Recall the definitions of $f_1, f_2, a$ in  (\ref{eqn_R}).
\subsection{CoRE Kernel, Type 2}

The second type of CoRE kernel perhaps  appears less intuitive than the first type:
\begin{align}
K_{C,2} = K_{C,2}(u,v) =  \rho \frac{\sqrt{f_1f_2}}{f_1+f_2-a} = \frac{\rho R }{a/\sqrt{f_1f_2}}
\end{align}
If the data are binary, then $K_{C,2}=R$. We will, later in the paper, also write $K_{C,2}$ as an expectation of inner product to confirm it is also positive definite.

\subsection{Kernel SVM Experiment}

\begin{figure}[h!]
\begin{center}
\mbox{
\includegraphics[width=1.75in]{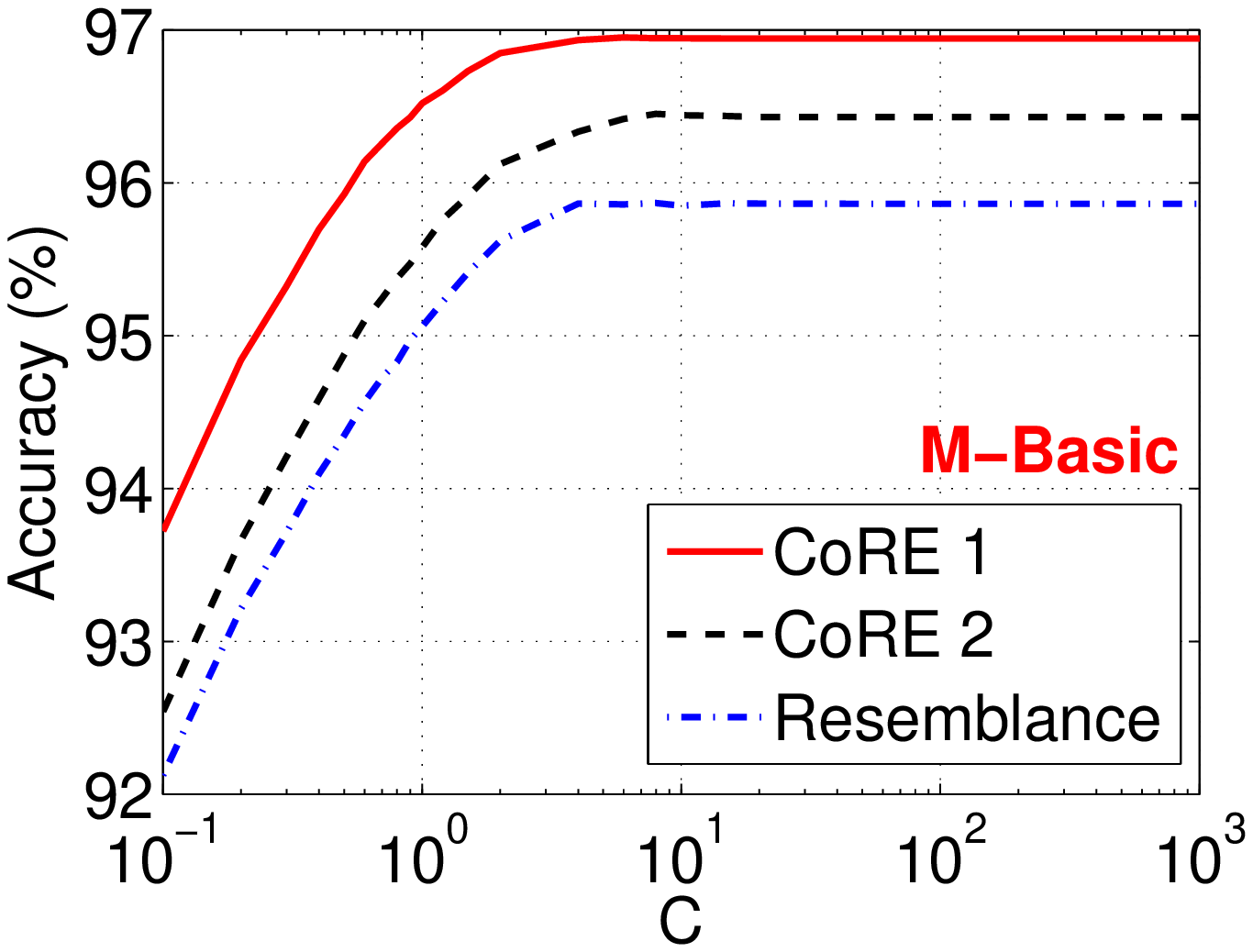}\hspace{-0.12in}
\includegraphics[width=1.75in]{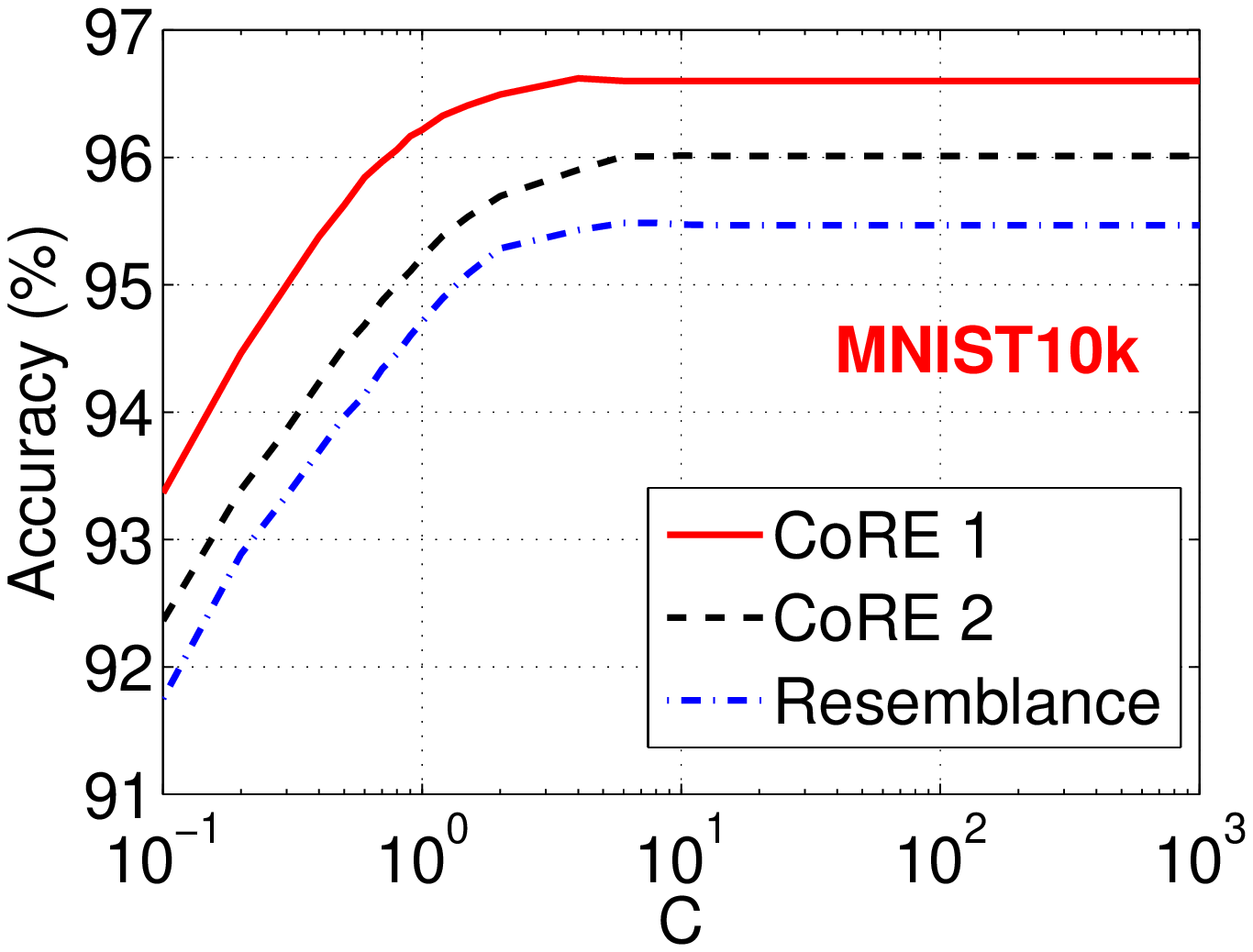}
}

\mbox{
\includegraphics[width=1.75in]{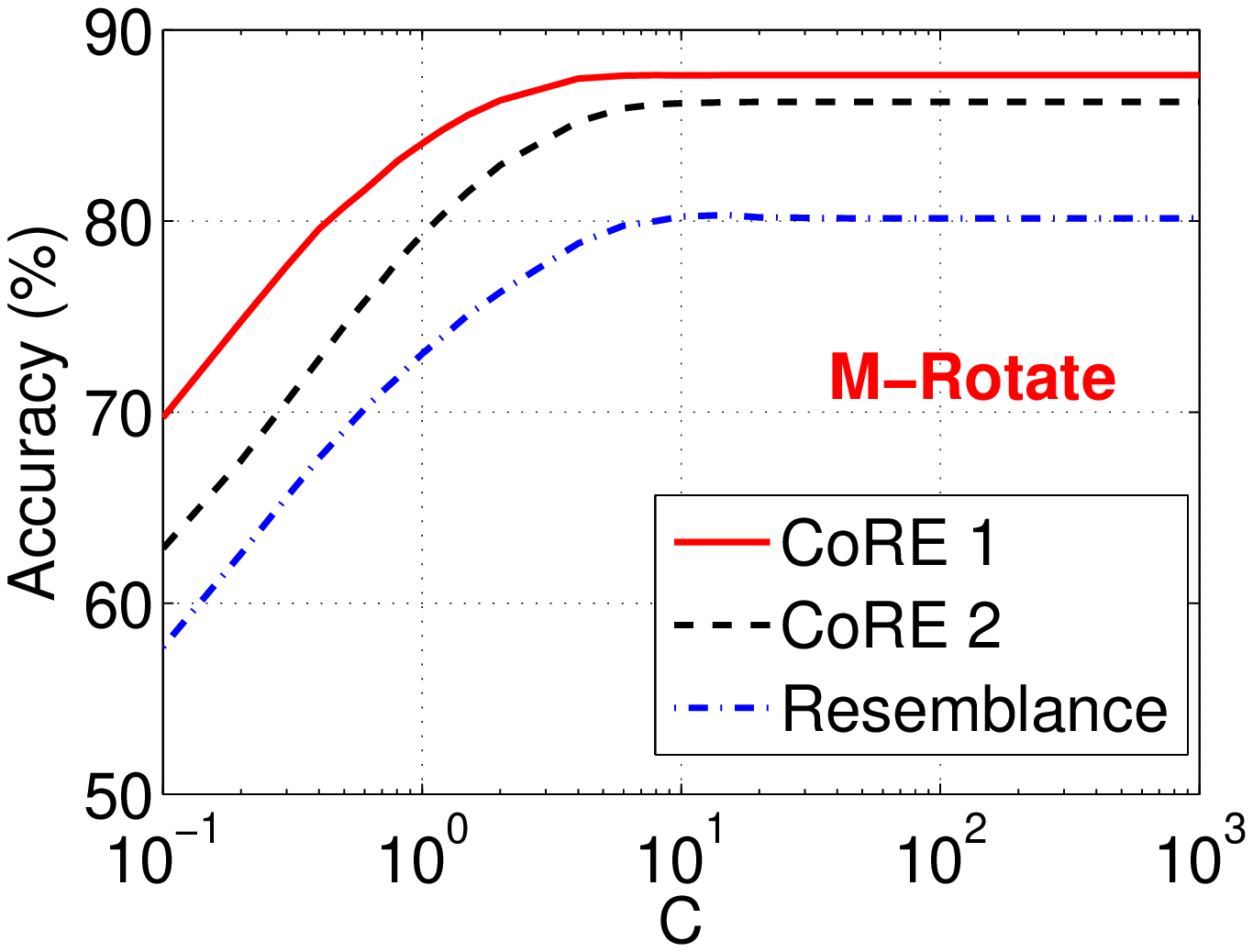}\hspace{-0.12in}
\includegraphics[width=1.75in]{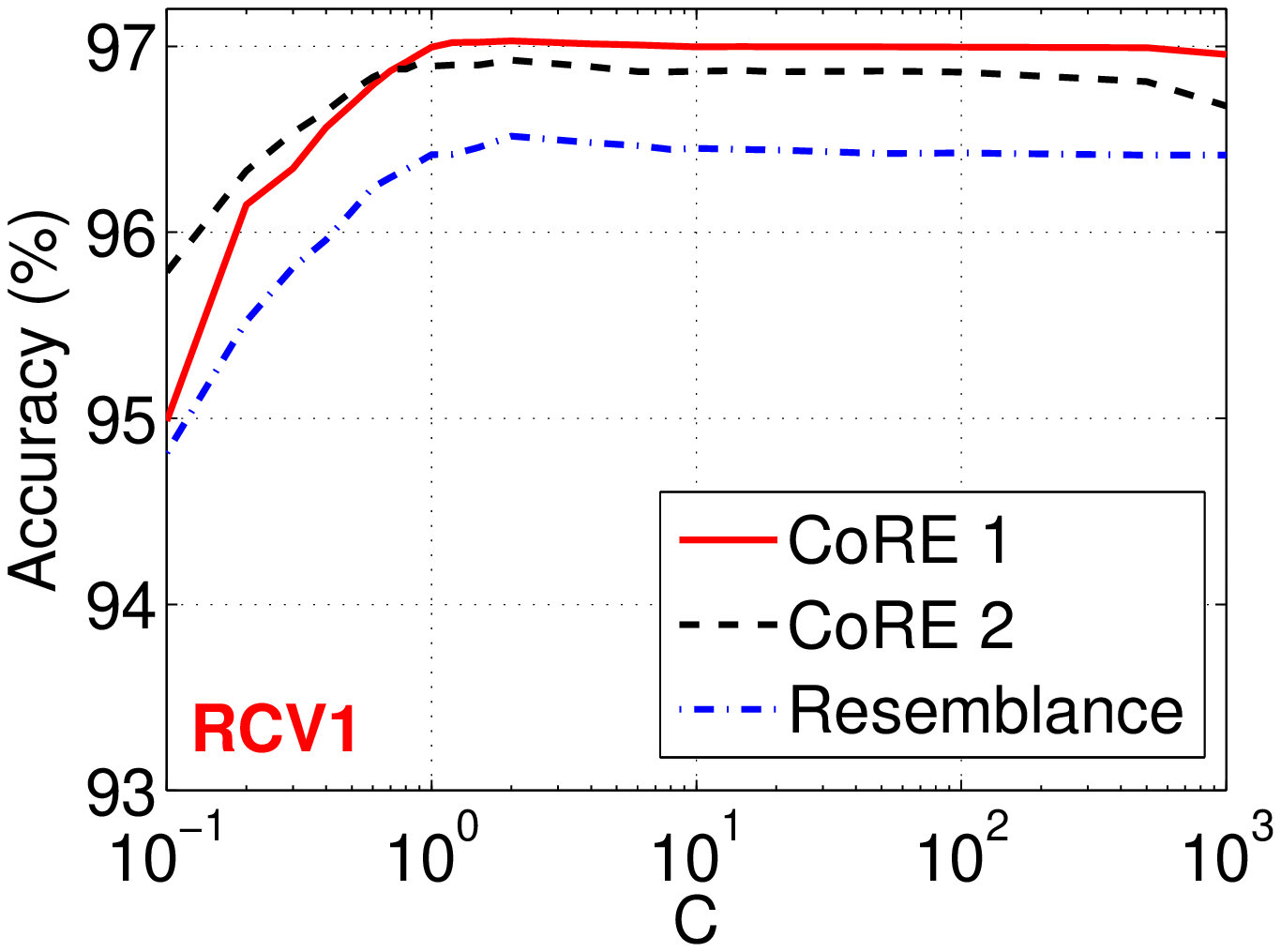}
}

\mbox{
\includegraphics[width=1.75in]{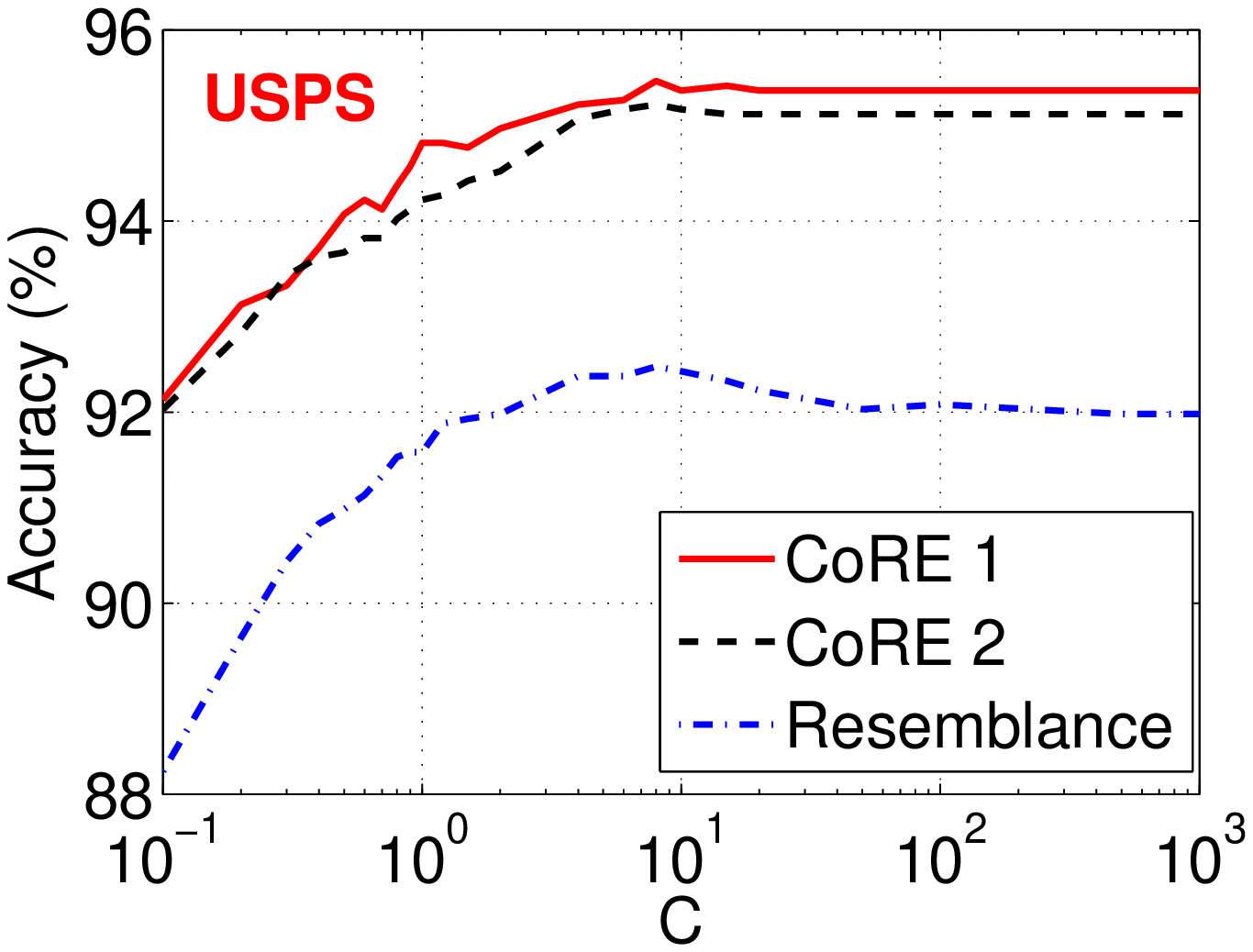}\hspace{-0.12in}
\includegraphics[width=1.75in]{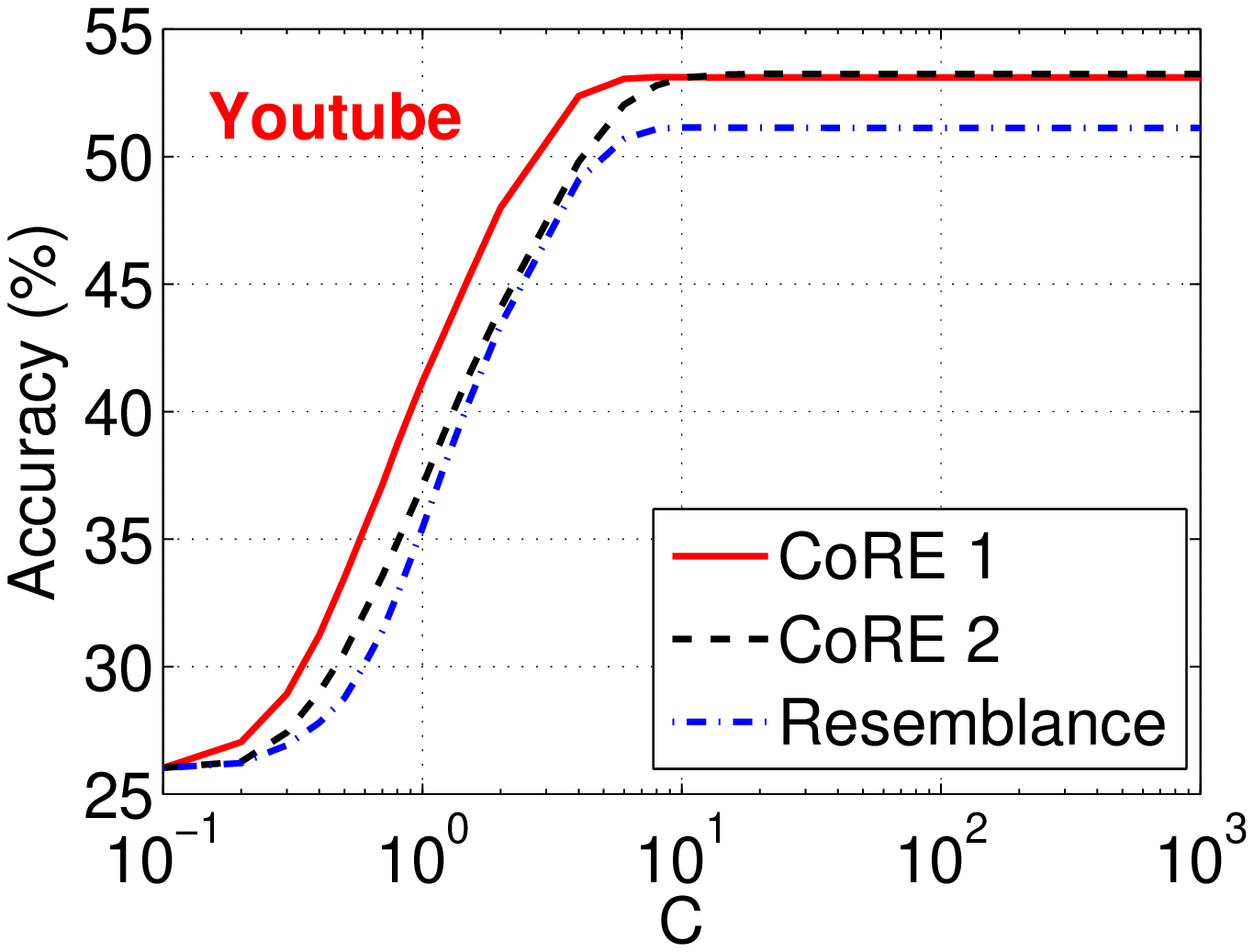}
}

\end{center}
\vspace{-0.2in}
\caption{Test classification accuracies using nonlinear kernel SVM and three types of kernels: CoRE Type 1, CoRE Type 2, and resemblance. We use LIBSVM pre-computed kernel functionality. Compared with the results of linear SVM in Figure~\ref{fig_linearSVM}, we can see CoRE kernels and  resemblance kernel perform better (or much better, especially M-Rotate).  The best results (highest points on the curves) are summarized in Table~\ref{tab_kernelSVM}. }\label{fig_kernelSVM}
\end{figure}

Figure~\ref{fig_kernelSVM} presents the classification accuracies on the same six datasets as in Figure~\ref{fig_linearSVM}, using nonlinear kernel SVM with three different kernels: CoRE Type 1, CoRE Type 2, and resemblance. We can see that resemblance (which only uses binary information of the data) does not perform as well as CoRE kernels.

\begin{table}[h!]
\caption{Best test classification accuracies  (in \%)  for  five different  kernels. The first two columns (i.e., ``linear'' and ``linear binary'') are already shown in Table~\ref{tab_linearSVM}.}
\begin{center}{\small
{\begin{tabular}{l c c c c c}
\hline \hline
Dataset        &Lin.    &Lin. Bin. &Res. & CoRE1 &CoRE2\\
\hline
M-Basic     &90.0   &88.9   &95.9   &\textbf{97.0}   &96.5 \\
MNIST10k      &90.0   &88.8   &95.5   &\textbf{96.6}   &96.0\\
M-Rotate    &48.0   &44.4   &80.3   &\textbf{87.6}   &86.2\\
RCV1       &96.3  &95.6   &96.5   &\textbf{97.0}  &96.9\\
USPS        &91.8   &87.4   &92.5  & \textbf{95.5}   &95.2\\
Youtube &47.6 &46.5 &51.1 &53.1 &\textbf{53.2}
\\\hline\hline
\end{tabular}}
}
\end{center}\label{tab_kernelSVM}

\end{table}

The best results are summarized in Table~\ref{tab_kernelSVM}. It is interesting to compare them  with   linear SVM results in Table~\ref{fig_linearSVM} and Figure~\ref{fig_linearSVM}. We can see that CoRE kernels perform  very well, without using additional tuning parameters. In fact, if we compare the best results in~\cite{Proc:Larochelle_ICML07,Proc:ABC_UAI10} (e.g., RBF SVM, abc-boosting, or deep learning) on MNIST10k, M-Rotate, and M-Basic, we will see that CoRE kernels (with no tuning parameters) can achieve the same (or  similar) performance.

We should mention that our experiments can be fairly easily reproduced because all datasets are public and we use standard SVM packages (LIBSVM and LIBLINEAR) without any modifications. We also provide the results for wide range of $C$ values in Figure~\ref{fig_linearSVM} and Figure~\ref{fig_kernelSVM}. Because we use pre-computed kernel functionality of LIBSVM (which consumes very substantial amount of memory to store the kernel matrices), we only experiment with  training data of moderate sizes, to ensure repeatability.

\subsection{Challenges with Nonlinear Kernel SVM}

\cite[Section 1.4.3]{Book:Bottou_07} mentioned three main computational issues of kernels  summarized as follows:
\begin{enumerate}
\item {\em Computing kernels is very expensive} \\
Computing kernels can account for more than half of the total computing time.
\item {\em Computing the full kernel matrix is wasteful}\\
This is because not all pairwise kernel values will be used during training.\vspace{-0.05in}
\item {\em  The kernel matrix does not fit in memory} \\
The cost of storing the full kernel matrix  in the memory is  $O(n^2)$, which is not realistic for most PCs even for  merely   $10^5$, while the industry has used training data with billions of examples.  Thus, kernel evaluations are often conducted {\em on the fly}, which means the computational cost is dominated by kernel evaluations.
\end{enumerate}
In fact, evaluating kernels on-demand would encounter another serious (and often common) issue if the datasets themselves are too big for the memory.

All these crucial issues motivate us to develop hashing algorithms to approximate CoRE kernels.

\subsection{Benefits of Hashing}

Our goal is to develop good hashing algorithms to (approximately) transform nonlinear kernels into linear kernels. Once we have the new data representations (i.e., the hashed data), we can use highly efficient batch or stochastic linear  methods for training SVM (or logistic regression)~\cite{Proc:Joachims_KDD06,Proc:Shalev-Shwartz_ICML07,URL:Bottou_SGD,Article:Fan_JMLR08}. Another benefit would be in the context of approximate near neighbor search because probabilistic hashing provides a (often good) strategy for space partitioning which will help reduce the search time (i.e., no need to scan all data points). Our proposed hashing methods can be modified to become an instance of {\em locality sensitive hashing (LSH)}~\cite{Proc:Indyk_STOC98} in the space of CoRE kernels.

In this study, we will present hashing algorithms for CoRE kernels based on standard random projection and minwise hashing methods. We first provide a review of these two methods.

\section{Review of Random Projections and Minwise Hashing}

\subsection{Random Projections}

Consider two vectors $u,v\in\mathbb{R}^{D}$. The idea of random projection is simple. We first generate a random vector of i.i.d. entries $r_i$, $i=1$ to $D$, and then compute the inner products as the hashed values:
\begin{align}\label{eqn_P}
P(u) =  \sum_{i=1}^D u_i r_i,\hspace{0.2in} P(v) = \sum_{i=1}^D v_i r_i
\end{align}
For the convenience of theoretical analysis, we adopt $r_i \sim N(0,1)$, which is a typical choice in the literature. Several variants of random projections like~\cite{Proc:Li_Hastie_Church_KDD06,Proc:Weinberger_ICML2009} are essentially equivalent, as analyzed in~\cite{Proc:HashLearning_NIPS11}.

We always assume  the data are normalized, i.e., $\sum_{i=1}^D u_i^2=\sum_{i=1}^D v_i^2 =1$. Note that computing the $l_2$ norms of all the data points only requires scanning the data once which is anyway needed during data collection/processing. For normalized data, it is known that $E\left[P(u)P(v)\right] = \rho$.   In order to estimate  $\rho$, we need to use $k$ random projections to generate $P_j(u), P_j(v), j = 1$ to $k$, and estimate $\rho$ by $\frac{1}{k}\sum_{j=1}^k P_j(u)P_j(v)$, which is also an inner product. This means we can directly use the projected data to build a linear classifier.

\subsection{Minwise Hashing}

The method of minwise hashing~\cite{Proc:Broder} is very popular for computing set  similarities, especially for industrial applications, for example,~\cite{Proc:Broder,Proc:Fetterly_WWW03,Proc:Henzinger_SIGIR06,Article:Urvoy08, Proc:Nitin_WSDM08,Proc:Chierichetti_KDD09,Proc:Gollapudi_WWW09,Proc:Najork_WSDM09,Proc:Buehrer_WSDM08}.

 Consider the space of the column numbers: $\Omega = \{1, 2, 3, ..., D\}$. We assume a random permutation $\pi: \Omega \longrightarrow\Omega$ and apply $\pi$ on the coordinates of both vectors $u$ and $v$. For example, consider $D=4$, $u = [0,\ 0.45,\ 0.89,\ 0]$ and $\pi: 1\rightarrow3,\ 2\rightarrow1,\ 3\rightarrow4,\ 4\rightarrow 2$. Then the permuted vector becomes $\pi(u) = [0.45,\ 0,\ 0,\ 0.89]$. In this example, the first nonzero column of $\pi(u)$ is 1, and the corresponding value of the coordinate is 0.45. For convenience, we introduce the following notation:
\begin{align}\label{eqn_L}
&L(u) = \text{location of  first nonzero entry of } \pi(u)\\\label{eqn_V}
&V(u) = \text{value of  first nonzero entry of } \pi(u)
\end{align}
In this example, we have $L(u) = 1$ and $V(u) = 0.45$.

The following  well-known {\em collision probability}
\begin{align}
\mathbf{Pr}\left(L(u) = L(v)\right) = R(u,v) = R
\end{align}
can be used  to estimate the resemblance $R$. To do so, we  need to generate $k$ permutations $\pi_j$, $j=1$ to $k$.

The proposed hashing algorithms for CoRE kernels combine  random projections and minwise hashing.
\section{Hashing  CoRE Kernels}\label{sec_hashing}

The goal is to develop unbiased estimators of $K_{C,1}$ and $K_{C,2}$ which can be written as  inner products.  We assume  that we have conducted random projections and minwise hashing $k$ times. In other words, for each data vector $u$, we have the hashed values $P_j(u)$, $L_j(u)$, $V_j(u)$,  $j=1$ to $k$. Recall the definitions of $P_j$, $L_j$, $V_j$ in (\ref{eqn_P}), (\ref{eqn_L}), and (\ref{eqn_V}), respectively.

\subsection{Hashing Type 1 CoRE Kernel}

Our proposed estimator of $K_{C,1}$ is
\begin{align}
\hat{K}_{C,1}(u,v) = \sum_{j=1}^k P_j(u)P_j(v)1\{L_j(u) = L_j(v)\}
\end{align}

\begin{theorem}\label{thm_Kc1}
\begin{align}
E\left(\hat{K}_{C,1}\right) = K_{C,1}
\end{align}
\begin{align}\label{eqn_VarKc1}
Var\left(\hat{K}_{C,1}\right) =\frac{1}{k}\left\{\left(1+2\rho^2\right)R -  \rho^2 R^2\right\}
\end{align}
\textbf{Proof:}\ \ See Appendix~\ref{app_thm_Kc1}.$\hfill\Box$
\end{theorem}

A simple argument can  show that $\hat{K}_{C,1}$ could be written as an inner product; and hence $K_{C,1}$ is positive definite. Although this fact is obvious since $K_{C,1}$ is a product of two positive definite kernels, we would like to present a constructive proof because the construction is basically the same procedure for expanding the hashed data before feeding them to an SVM solver.

 Recall, $L_j$ is the location of the first nonzero after minwise hashing.  Basically, we can view $L_j(u)$  equivalently as a vector of length $D$ whose coordinates are all zero except the $L_j(u)$-th coordinate. The value of the only nonzero coordinate will be $P_j(u)$. For example, suppose $D=4$, $L_j(u) = 2$, $P_j(u)=0.1$. Then the equivalent vector would be $[0,\ 0.1,\ 0,\ 0]$. This way, we can write $\hat{K}_{C,1}$ as an inner product of two $D$-dimensional sparse vectors.

Note that the  input data format of standard SVM packages is  the sparse format. For linear SVM, the  cost is essentially determined by the number of nonzeros (in this case, $k$), not much to do with the dimensionality (unless it is too high). If $D$ is too high, then we can adopt the standard trick of {\em $b$-bit minwise hashing}~\cite{Proc:Li_Konig_WWW10,Proc:HashLearning_NIPS11} by only using the lowest $b$ bits of $L_j(u)$.

\subsection{Hashing Type 2 CoRE Kernel}

Our second proposal is
\begin{align}
\hat{K}_{C,2} = \frac{\sqrt{f_1f_2}}{k}\sum_{j=1}^k V_j(u) V_j(v)1\{L_j(u) = L_j(v)\}
\end{align}
Recall that we always assume  the  data ($u$, $v$) are normalized. For example, if the data are binary, then we have $u_i = \frac{1}{\sqrt{f_1}}$, $v_i = \frac{1}{\sqrt{f_2}}$. Hence the values $V_j(u)$ and $V_j(v)$ are small  (and we need the term $\sqrt{f_1f_2}$).
\begin{theorem}\label{thm_Kc2}
\begin{align}
E\left(\hat{K}_{C,2}\right) = K_{C,2}
\end{align}
\begin{align}\label{eqn_VarKc2}
&Var\left(\hat{K}_{C,2}\right)\\\notag
=& \frac{1}{k}\frac{f_1f_2}{f_1+f_2-a}\left(\sum_{i=1}^D u_i^2v_i^2-\frac{\left(\sum_{i=1}^D u_iv_i\right)^2}{(f_1+f_2-a)}\right)
\end{align}
\textbf{Proof:}\ \ See Appendix~\ref{app_thm_Kc2}.$\hfill\Box$
\end{theorem}

Once we understand how to express $\hat{K}_{C,1}$ as an inner product, it should be easy to see that $\hat{K}_{C,2}$ can also be written as an inner product. Again, suppose $D=4$, $L_j(u) = 2$, and $V_j(u)=0.05$. We can consider an equivalent vector $[0,\ 0.05f_1, 0,\ 0]$. In other words, the difference between $\hat{K}_{C,1}$ and $\hat{K}_{C,2}$ is what value we should put in the nonzero location. One advantage of $\hat{K}_{C,2}$ is that it only requires the permutations and thus eliminates the cost of random projections.

As expected, the variance of $\hat{K}_{C,2}$ would be  large if the data are heavy-tailed. However, when the data are binary or  appropriately normalized (e.g.,   TF-IDF), $Var\left(\hat{K}_{C,2}\right)$ is actually quite small. For example, when the data are binary, i.e., $u_i = \frac{1}{\sqrt{f_1}}$, $v_i=\frac{1}{\sqrt{f_2}}$, we have
$Var\left(\hat{K}_{C,2}\right)= \frac{1}{k}\left(R-R^2\right)$, which is (considerably) smaller than  $Var\left(\hat{K}_{C,1}\right) =\frac{1}{k}\left\{\left(1+2\rho^2\right)R -  \rho^2 R^2\right\}$.

\subsection{Experiment for Validation}

\begin{figure}[h!]
\begin{center}

\mbox{
\includegraphics[width=1.75in]{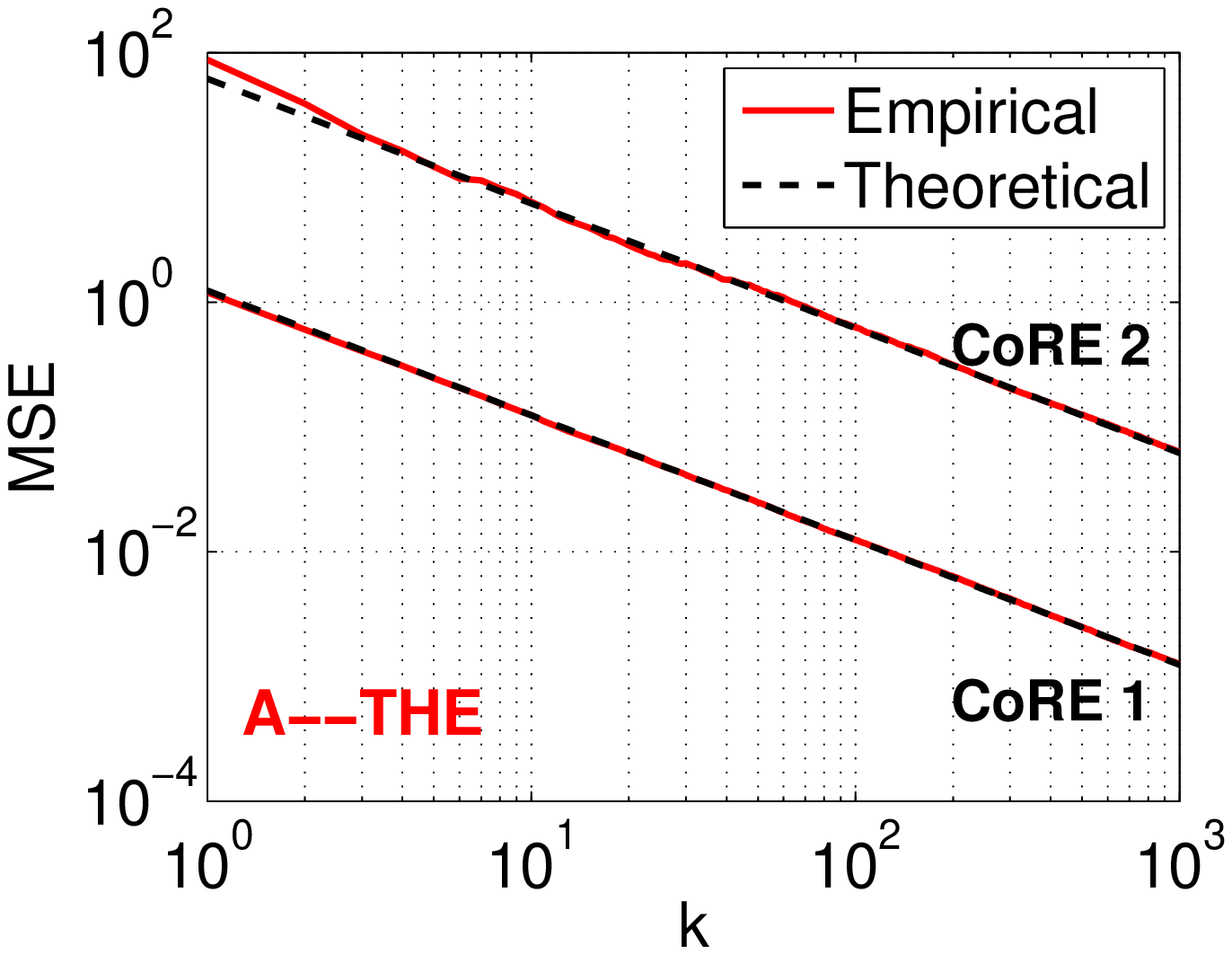}\hspace{-0.12in}
\includegraphics[width=1.75in]{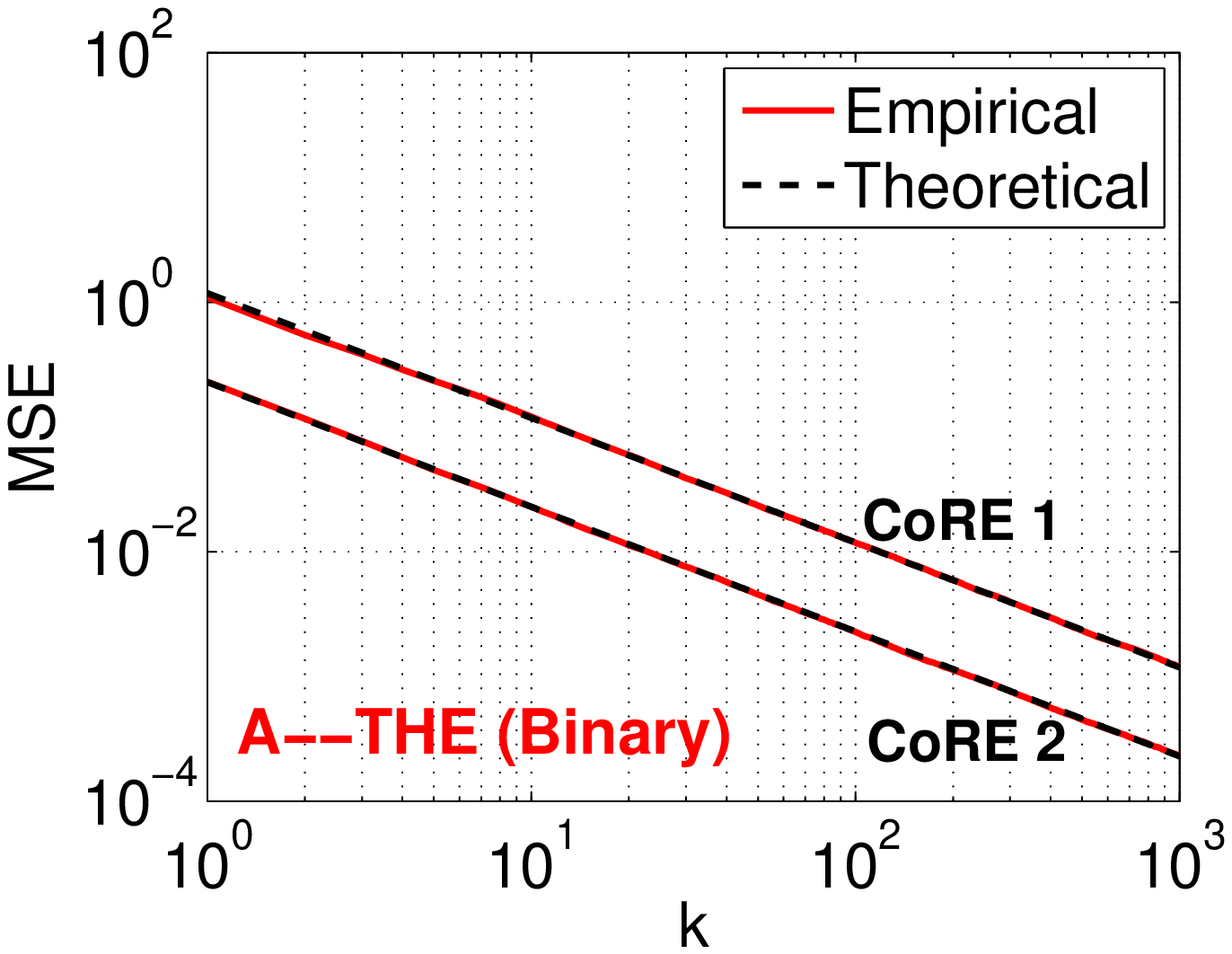}
}
\mbox{
\includegraphics[width=1.75in]{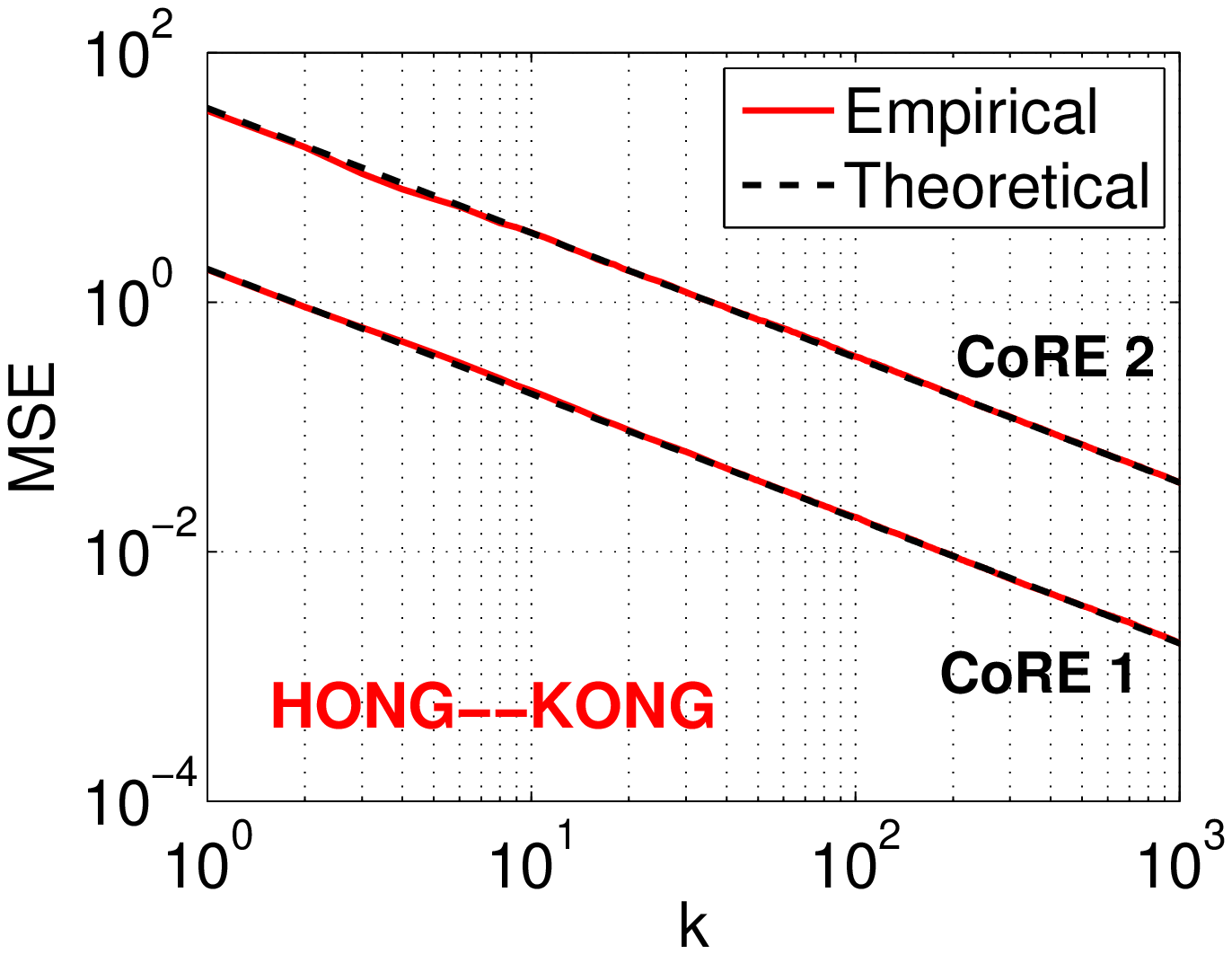}\hspace{-0.12in}
\includegraphics[width=1.75in]{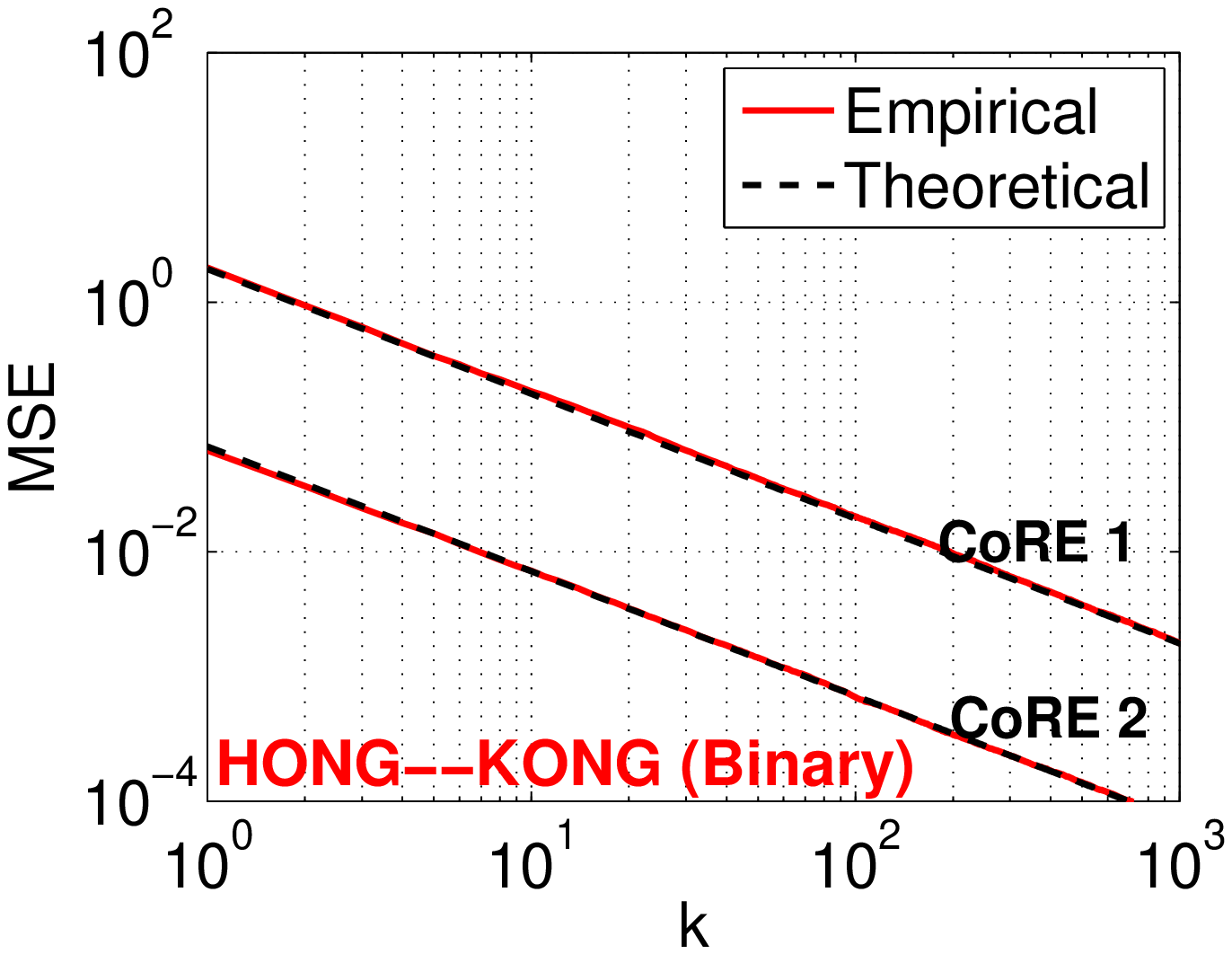}
}

\end{center}
\vspace{-0.2in}
\caption{Mean square errors (MSE = Var + Bias$^2$) on two pairs of word vectors for validating Theorems~\ref{thm_Kc1} and~\ref{thm_Kc2}. The empirical MSEs (solid curves) essentially overlap the theoretical variances (dashed curves), (\ref{eqn_VarKc1}) and (\ref{eqn_VarKc2}). When using the raw counts (left panels), the MSEs of $\hat{K}_{C,2}$ is significantly higher than the MSEs of $\hat{K}_{C,1}$. However, when using binarized data (right panels), the MSEs of $\hat{K}_{C,2}$ become noticeably smaller.}\label{fig_MSE}
\end{figure}

To validate the theoretical results in Theorem~\ref{thm_Kc1} and Theorem~\ref{thm_Kc2}, we provide a set of experiments in Figure~\ref{fig_MSE}. Two pairs of word vectors are selected: ``A--THE'' and ``HONG--KONG'', from a chuck of web crawls. For example, the vector ``HONG'' is a vector whose $i$-th entry is the number of occurrences of the word ``HONG'' in the $i$-th document. For each pair, we apply the two proposed hashing algorithms on the two corresponding vectors to estimate $K_{C,1}$ and $K_{C,2}$. With sufficient repetitions, we can empirically compute the mean square errors (MSE = Var + Bias$^2$), which should match the theoretical variances if the estimators are indeed unbiased and  the variance formulas, (\ref{eqn_VarKc1}) and (\ref{eqn_VarKc2}), are correct.

The number of word occurrences is a  typical example of highly heavy-tailed data. Usually when  text data are used in machine learning tasks, they have to be appropriately weighted (e.g., TF-IDF) or simply binarized. Figure~\ref{fig_MSE} presents the results on the original raw data as well as binarized data, to verify the formulas for these two extreme cases, for $k=1$ to 1000.

Indeed, the plots show that the empirical MSEs essentially overlap the theoretical variances. In addition, the MSEs of $\hat{K}_{C,2}$ is significantly larger than the MSEs of $\hat{K}_{C,1}$ on the raw data, as expected. Once the data are binarized, the MSEs of $\hat{K}_{C,2}$ become smaller.

\section{Hashing CoRE Kernels for SVM}

In this section, we provide a set of experiment for using  the  hashed data  as input for a linear SVM solver (LIBLINEAR). Our goal is to  approximate the performance of (nonlinear) CoRE kernels with linear kernels. In Section~\ref{sec_hashing}, we  have explained how to express the estimators $\hat{K}_{C,1}$ and $\hat{K}_{C,2}$ as inner products by expanding the hashed data. With $k$ permutations and $k$ random projections, the number of nonzeros of the expanded data is precisely $k$. To reduce the dimensionality, we use only the lowest $b$ bits of the locations. In this study, we experiment with $b=1$, 2, 4, 8.

\begin{figure*}[h!]
\begin{center}
\mbox{
\includegraphics[width=2.3in]{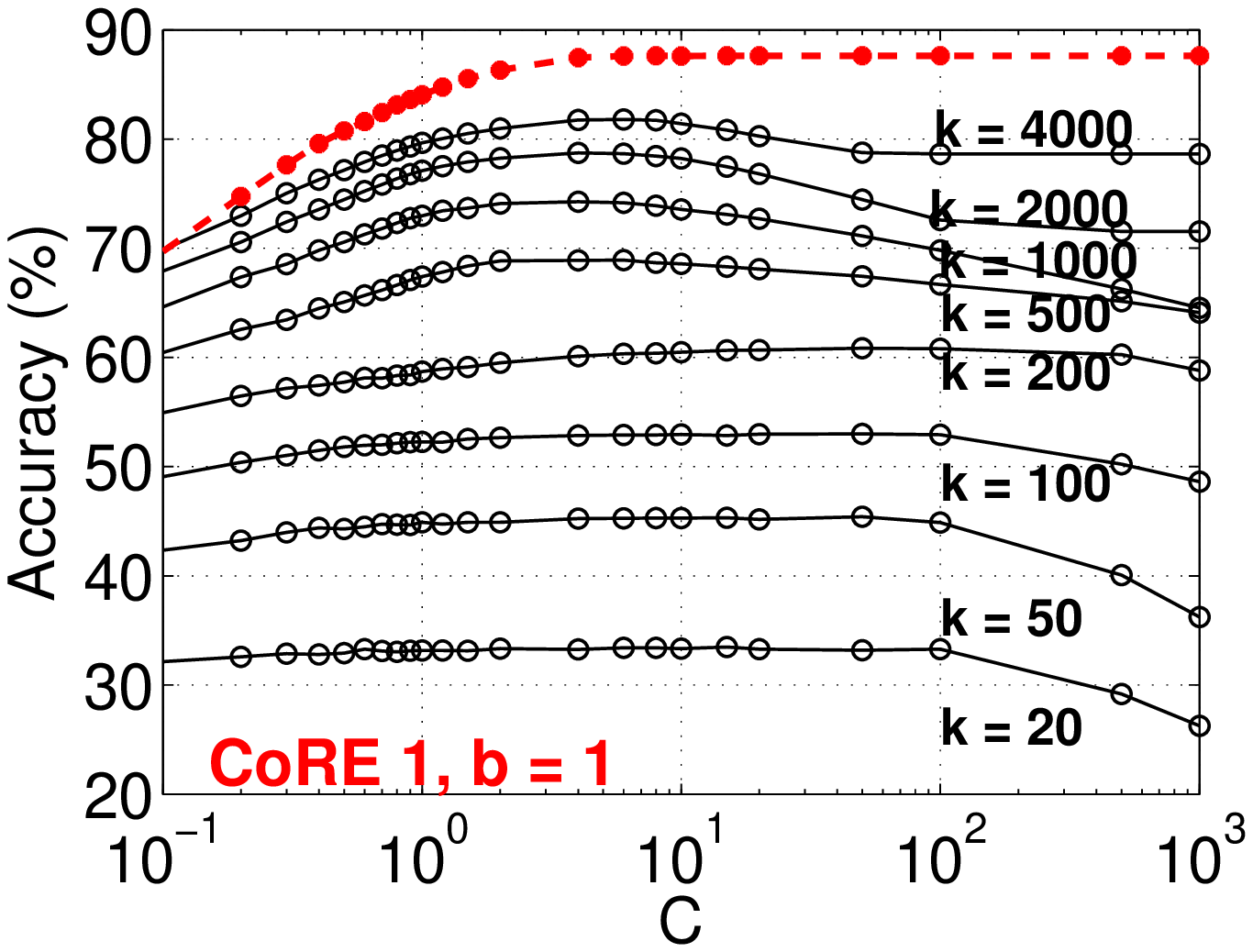}\hspace{-0.12in}
\includegraphics[width=2.3in]{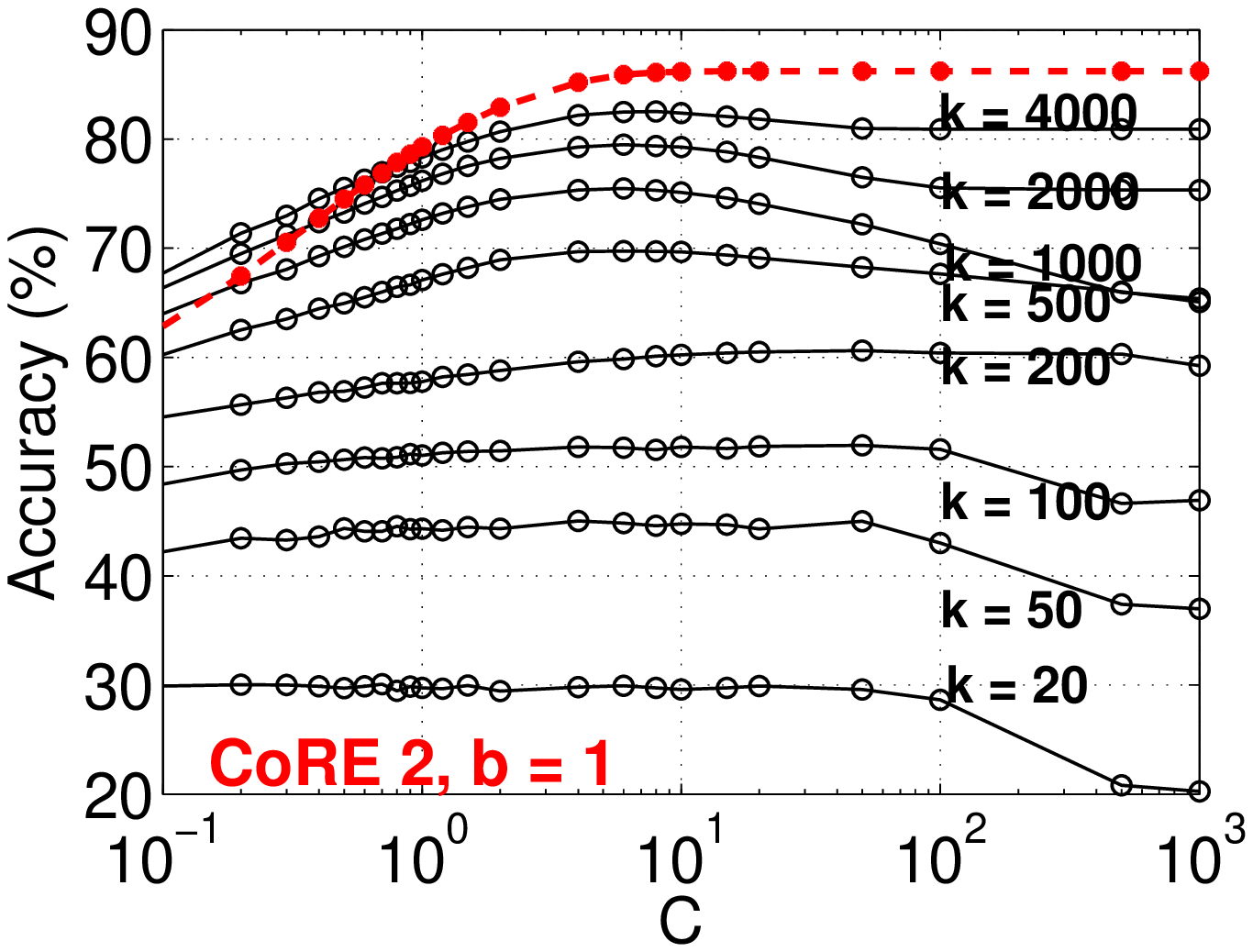}\hspace{-0.12in}
\includegraphics[width=2.3in]{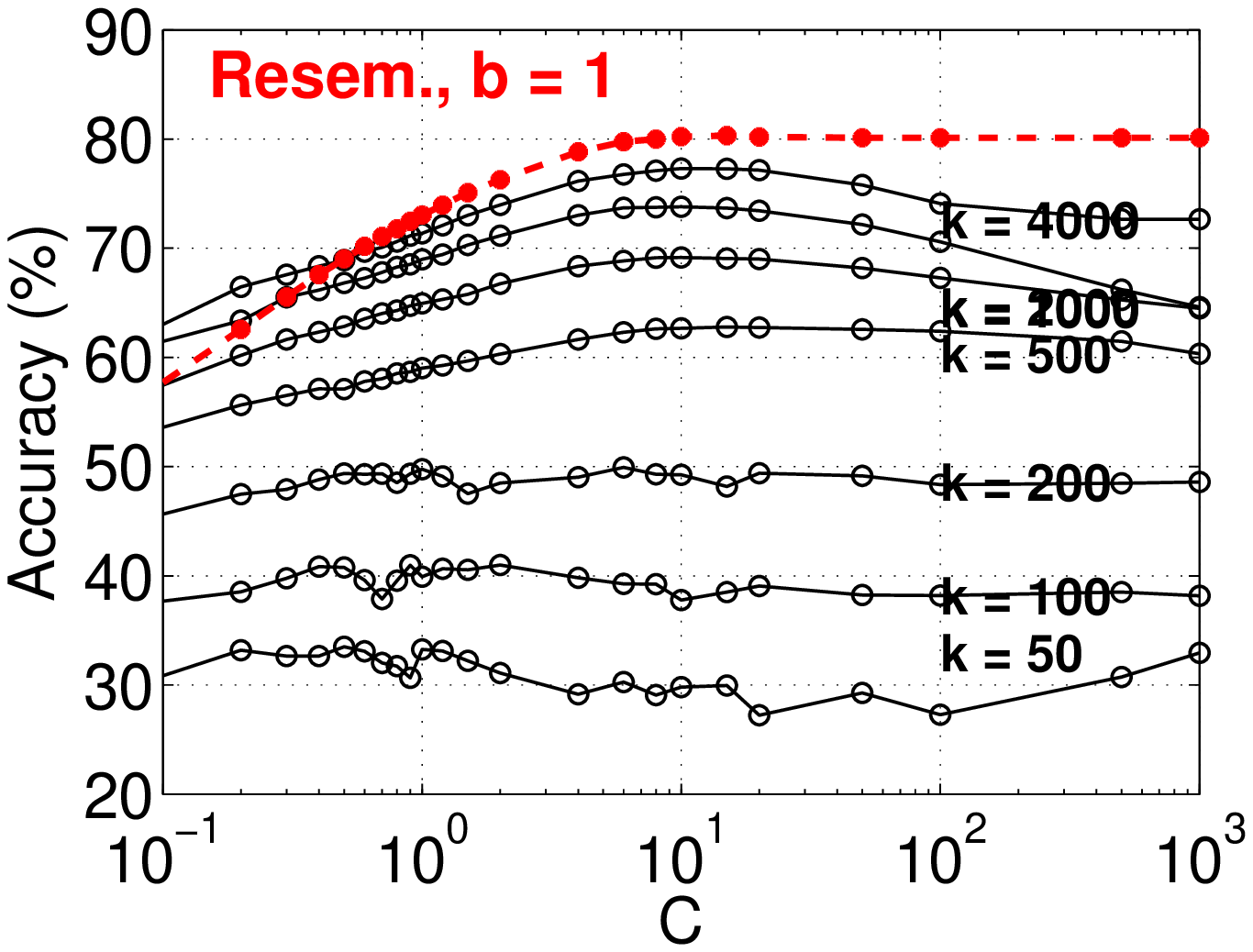}
}
\mbox{
\includegraphics[width=2.3in]{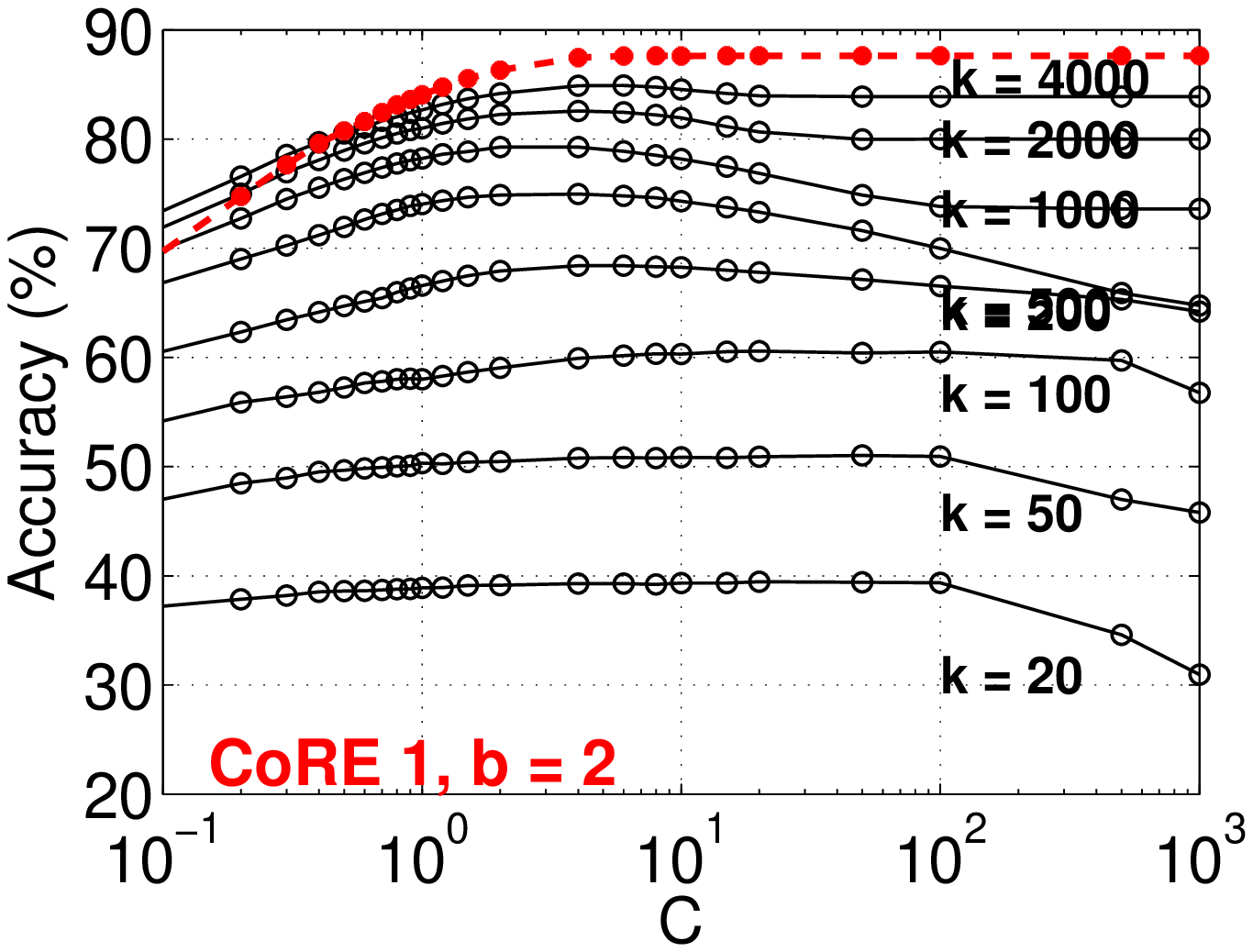}\hspace{-0.12in}
\includegraphics[width=2.3in]{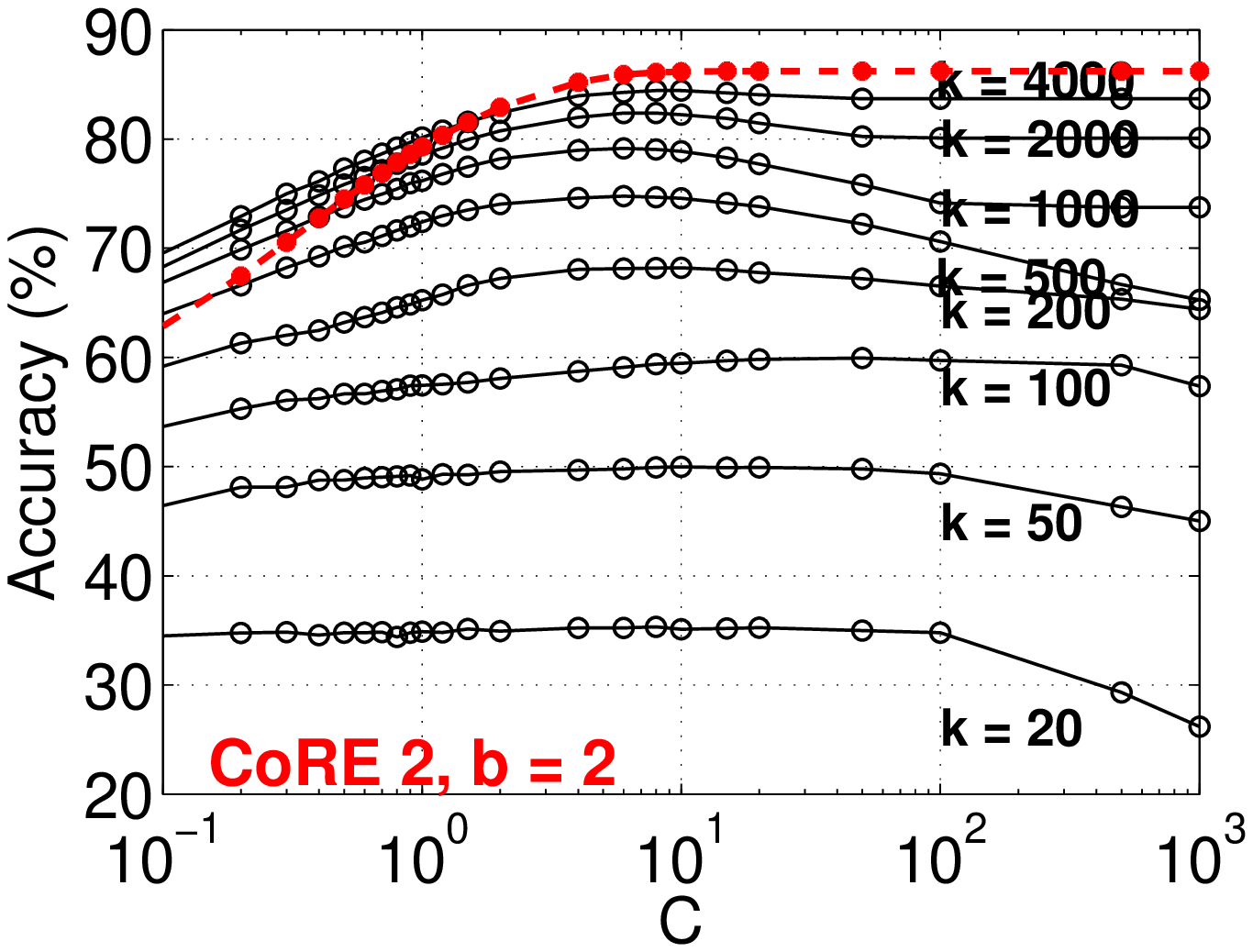}\hspace{-0.12in}
\includegraphics[width=2.3in]{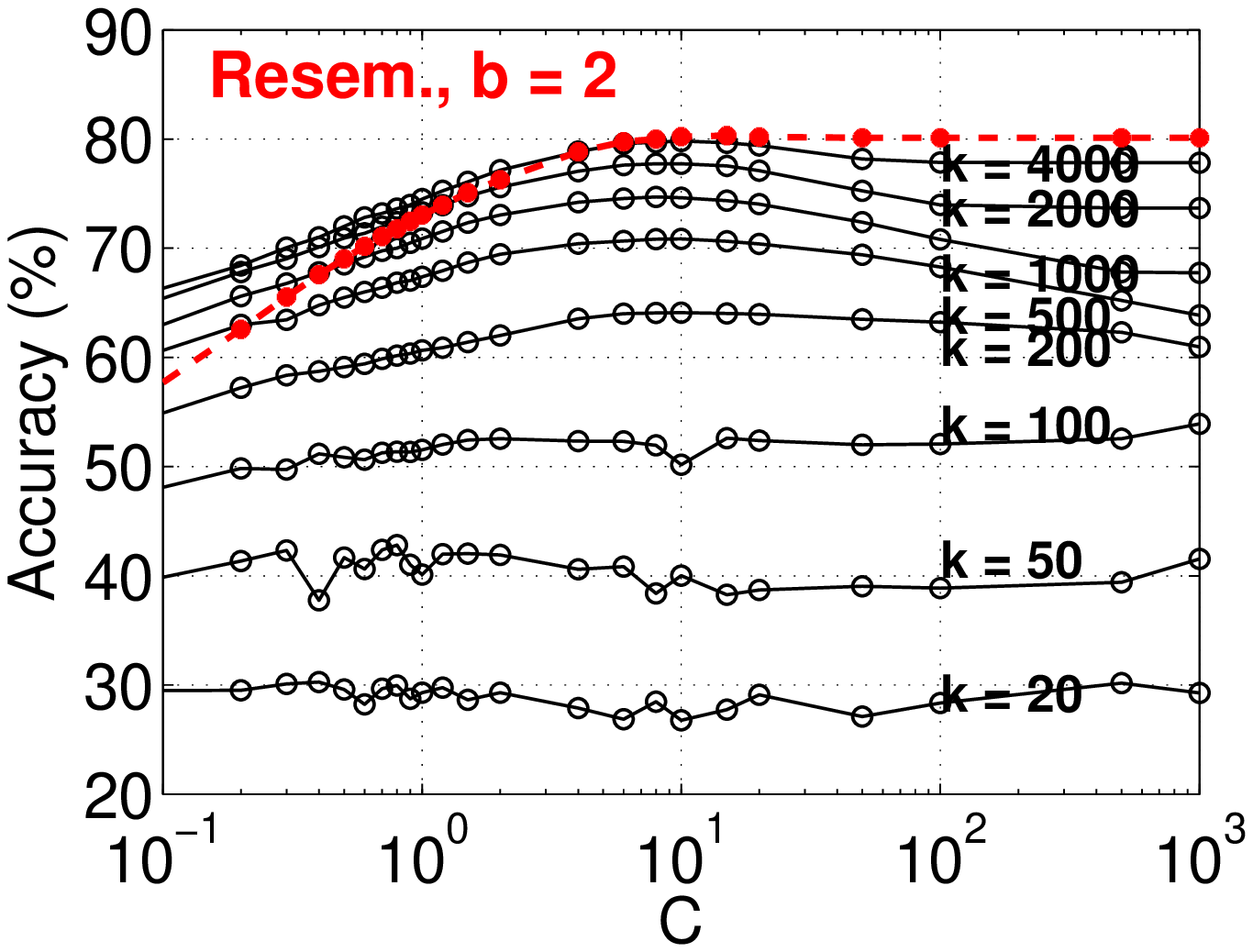}
}

\mbox{
\includegraphics[width=2.3in]{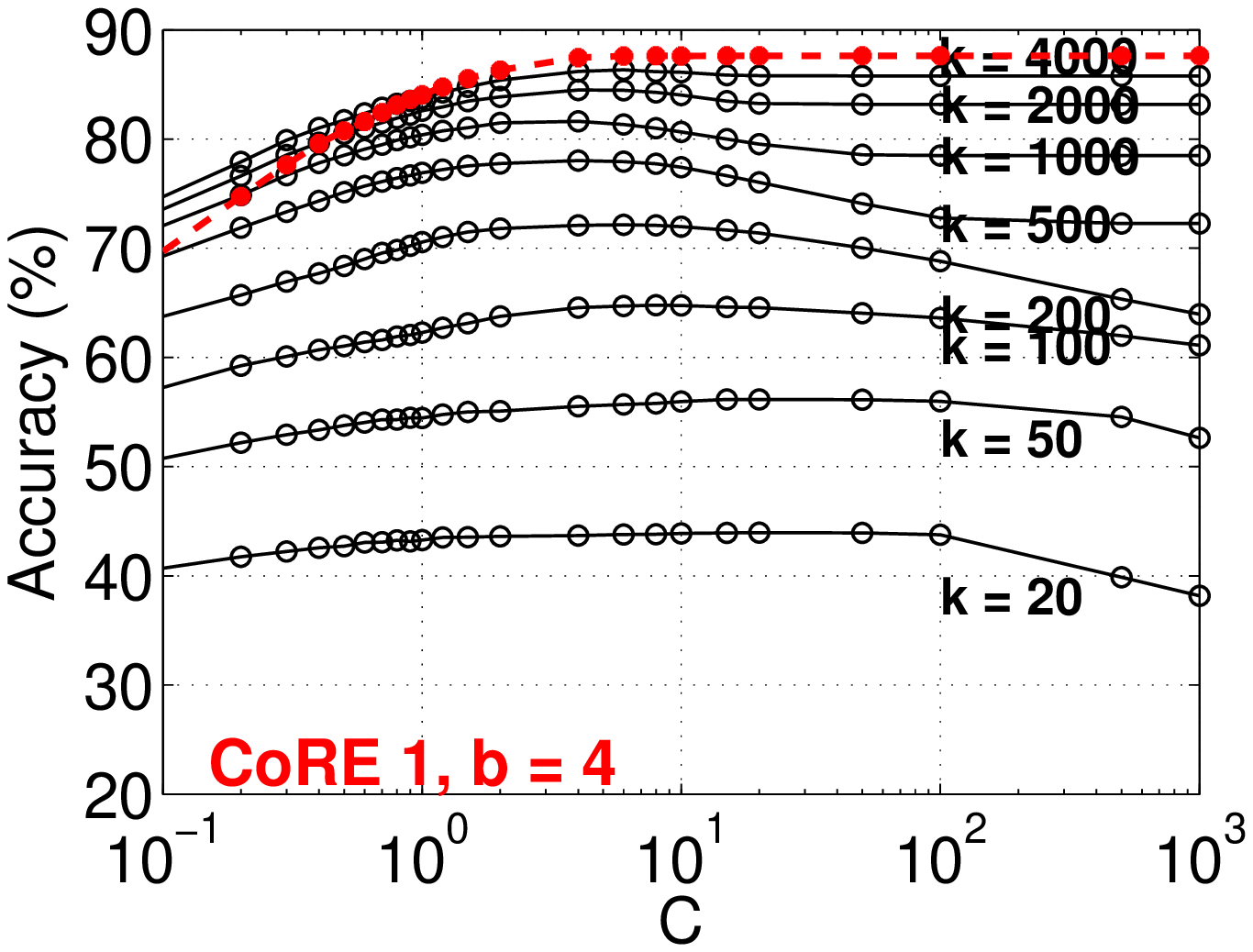}\hspace{-0.12in}
\includegraphics[width=2.3in]{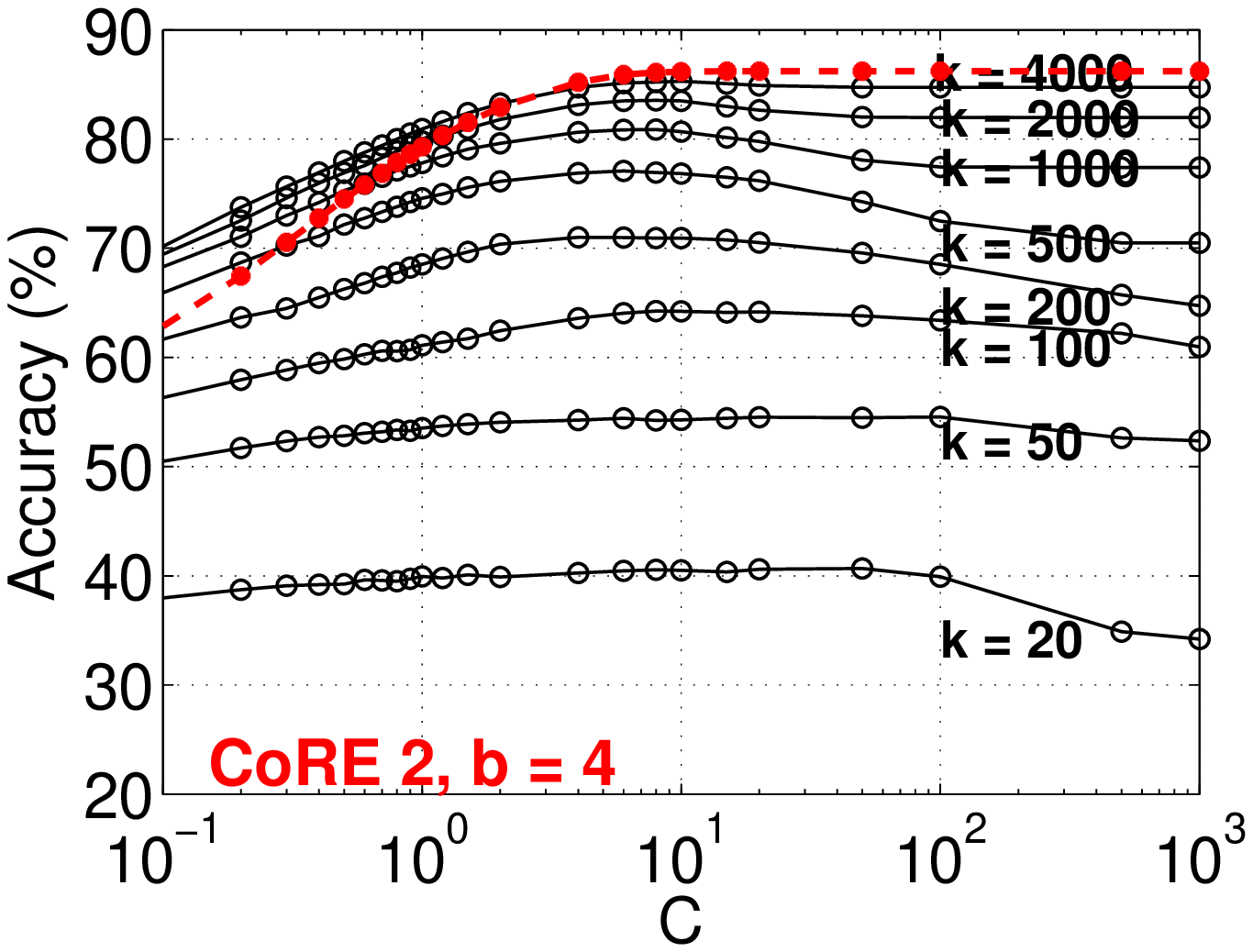}\hspace{-0.12in}
\includegraphics[width=2.3in]{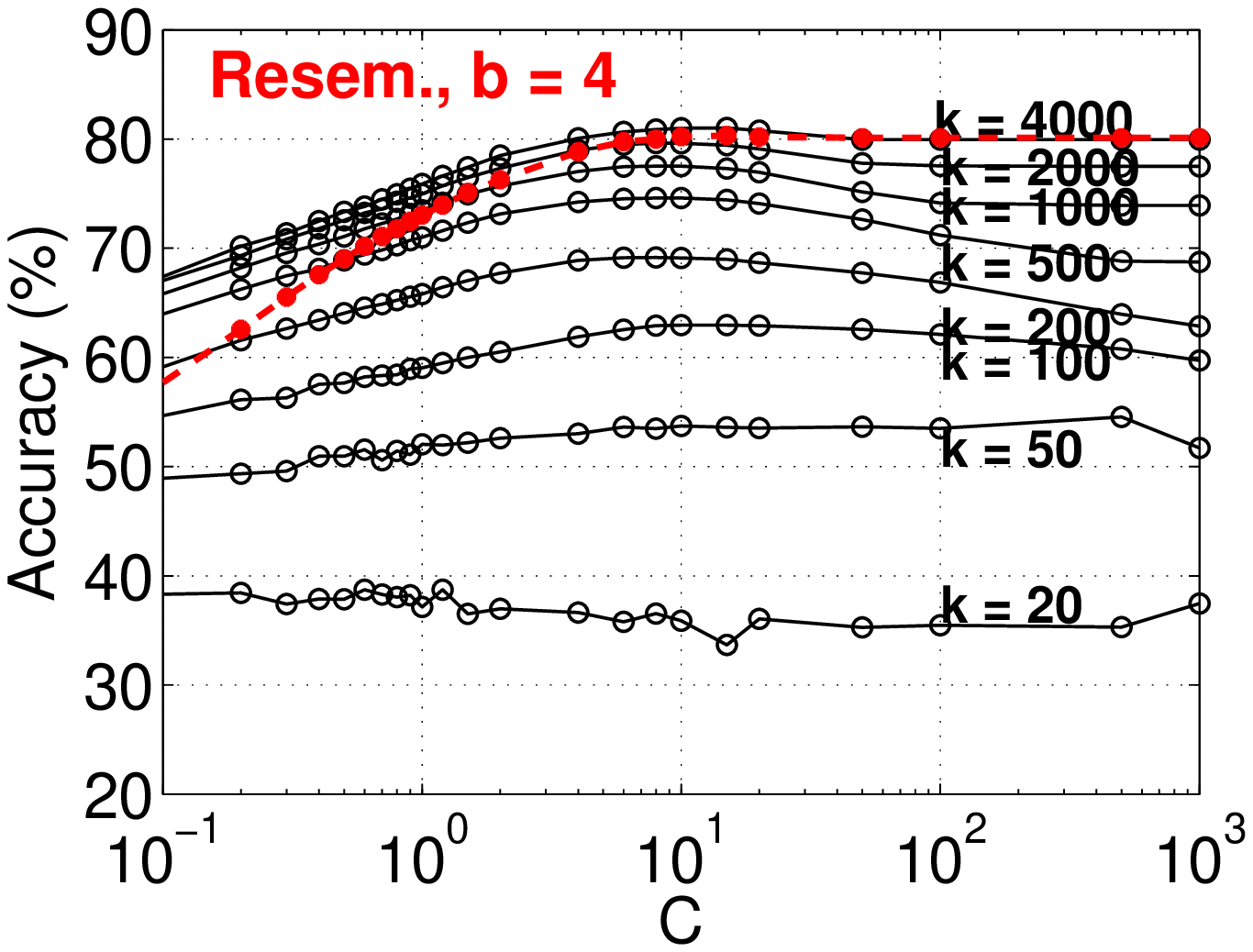}
}
\mbox{
\includegraphics[width=2.3in]{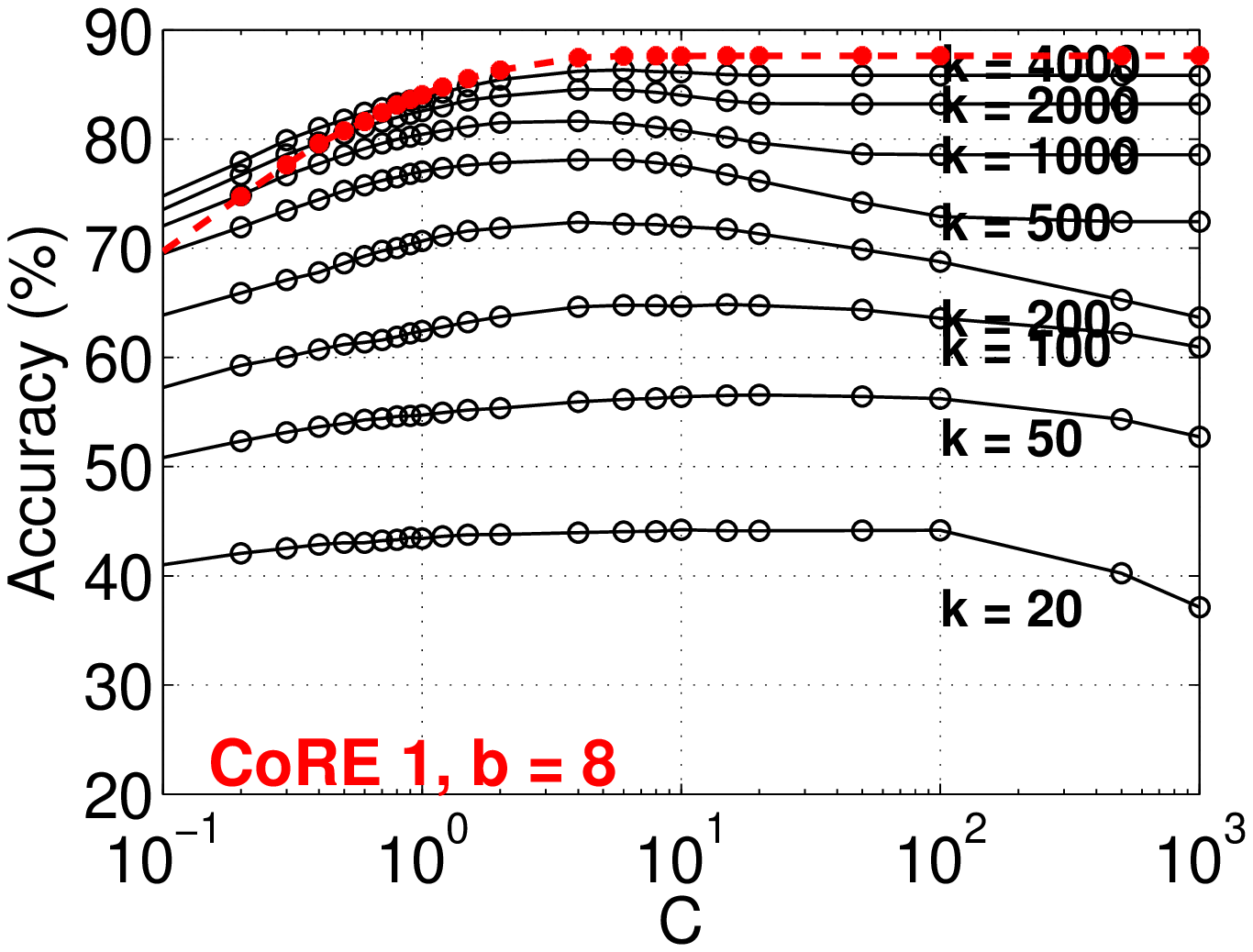}\hspace{-0.12in}
\includegraphics[width=2.3in]{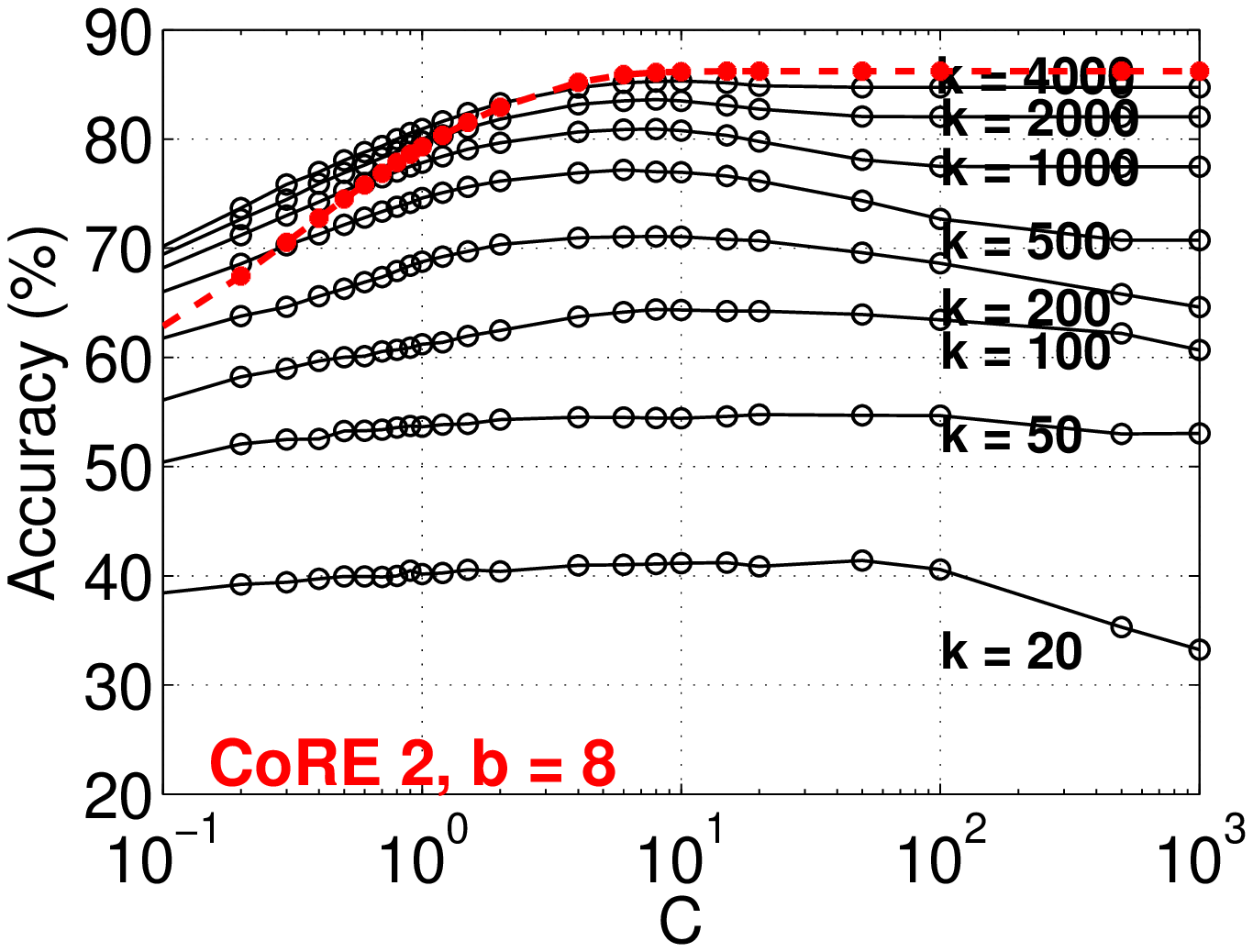}\hspace{-0.12in}
\includegraphics[width=2.3in]{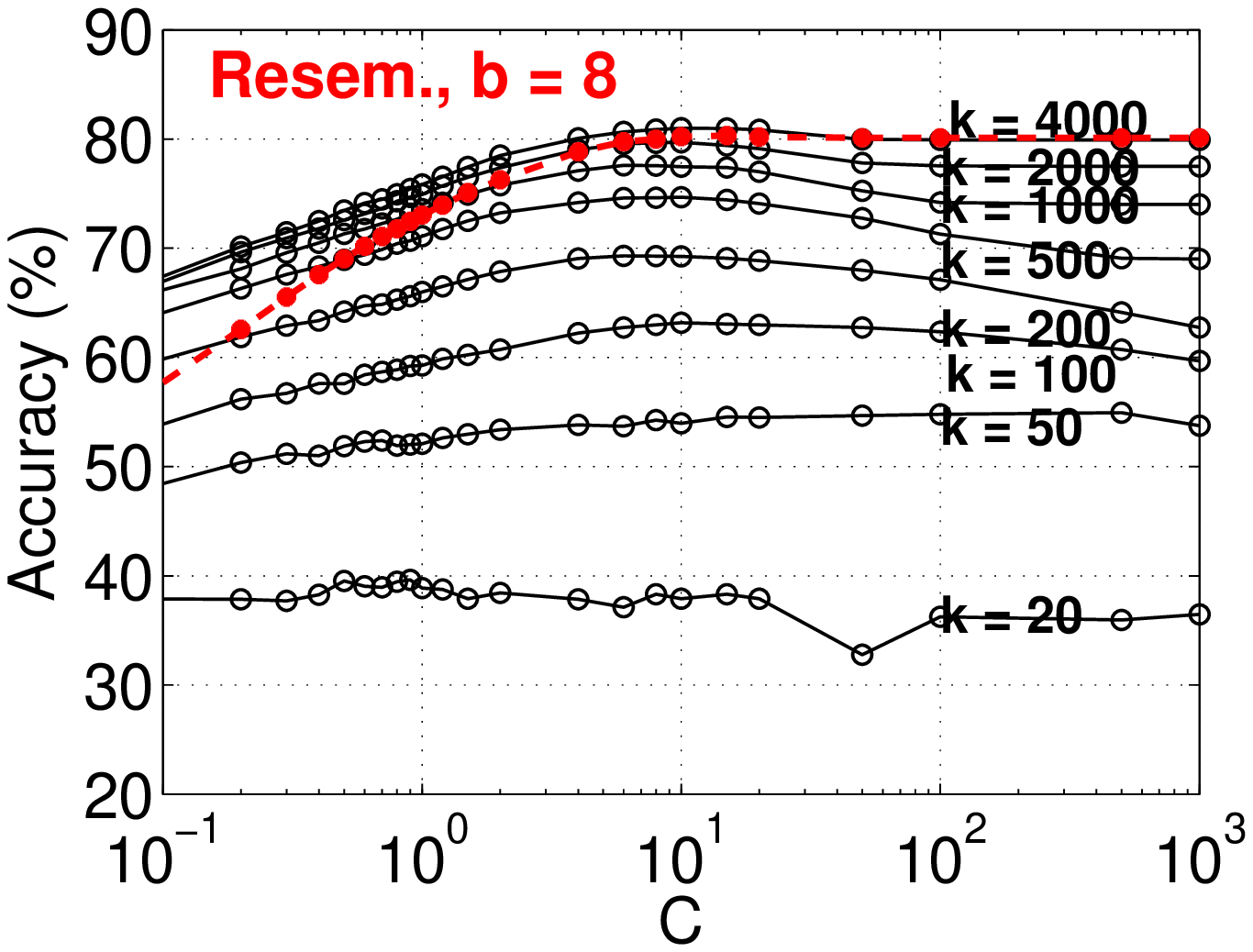}
}
\end{center}
\vspace{-0.2in}
\caption{Classification accuracies on the M-Rotate dataset using our proposed hashing methods and  linear SVM (LIBLINEAR). The red (if color is available) dot curves are the results of  kernel SVM  on the original data (i.e., the same curves from Figure~\ref{fig_kernelSVM}), using Type 1  CoRE kernel (left panels), Type 2  CoRE kernel (middle panels), and resemblance kernel (right panels), respectively. We apply both $b$-bit minwise hashing (with $b=1, 2, 4, 8$) and random projections $k$ times  and feed the (expanded) hashed data to linear SVM. }\label{fig_MRotateAcc}
\end{figure*}

Figure~\ref{fig_MRotateAcc} presents the results on the \textbf{M-Rotate} dataset. As shown in Figure~\ref{fig_linearSVM} and Table~\ref{tab_linearSVM}, using the linear kernel can only achieve an accuracy of $48\%$. This means, if we use random projections (or variants, e.g., ~\cite{Proc:Li_Hastie_Church_KDD06,Proc:Weinberger_ICML2009}), which approximate inner products, then the most we can achieve would be about $48\%$. For this dataset, the performance of CoRE kernels (and resemblance kernel) is astonishing, as shown in Figure~\ref{fig_kernelSVM} and Table~\ref{tab_kernelSVM}. Thus, we choose this dataset to demonstrate that our proposed hashing algorithms combined with linear SVM can also achieve the performance of (nonlinear) CoRE kernels.\\

To better explain the procedure, we use the same examples as in Section~\ref{sec_hashing}. Suppose  we apply $k$ minwise hashing and $k$ random projections on the data and we consider without loss of generality the data vector $u$.  For the $j$-th projection and $j$-th minwise hashing, suppose $L_j(u) = 2, V_j(u) = 0.05, P_j(u)  = 0.1$.  Recall $L_j$  and $V_j$ are, respectively, the location and the value of the first nonzero entry after minwise hashing. $P_j$ is projected value obtained from random projection.

In order to use linear SVM to approximate kernel SVM with Type 1 CoRE kernel, we expand the $j$-th hashed data as  a vector $[0,\ 0.1,\ 0, \ 0]$ if $b=2$, or $[0,\ 0.1]$ if $b=1$. We then concatenate $k$ such vectors to form a vector of length $2^b\times k$ (with exactly $k$ nonzeros). Before we feed the expanded hashed data to LIBLINEAR, we always normalize the vectors to have unit norm.  The experimental results are presented in the right panels of Figure~\ref{fig_MRotateAcc}.

To approximate Type 2 CoRE kernel, we expand the $j$-th hashed data of $u$ as $[0,\ 0.05f_1,\ 0,\ 0]$ if $b=2$, or $[0,\ 0.05f_1]$ if $b=1$, where $f_1$ is the number of nonzero entries in the original data vector $u$. Again, we concatenate $k$ such vectors. The experimental results are presented in the middle panels of Figure~\ref{fig_MRotateAcc}.

To approximate resemblance kernel, we expand the $j$-th hashed data of $u$ as $[0,\ 1,\ 0,\ 0]$ if $b=2$ or $[0,\ 1]$ if $b=1$ and we concatenate $k$ such vectors.

The results in Figure~\ref{fig_MRotateAcc} are exciting because linear SVM on the original data can only achieve an accuracy of $48\%$. Our proposed hashing methods + linear SVM can achieve $>86\%$. Using only the original $b$-bit minwise hashing, the accuracy can still reach about $80\%$. Again, we should  mention that other hashing algorithms which aim at approximating the inner product (such as random projections and variants) can at most achieve the same result as using linear SVM on the original data.

\section{Discussions}

There is a line of work called {\em Conditional Random Sampling (CRS)}~\cite{Proc:Li_Church_Hastie_NIPS06} which was also designed for sparse non-binary data. Basically, the idea of CRS is to keep the first $k$  nonzero entries after applying one permutation on the data. \cite{Proc:Li_Church_Hastie_NIPS06}  developed the trick to construct an (essentially) equivalent random sample for each pair. Although CRS is applicable to non-binary data, it is not suitable for training linear SVM (or other applications which require the input data to be in a metric space), because the hashed data of CRS are not appropriately aligned, unlike our method. 

Table~\ref{tab_kernelSVM} shows that Type 1 CoRE kernel can often achieve better results than Type 2 CoRE kernel, in some cases quite substantially. This helps justify the need for developing hashing methods for Type 1 CoRE kernel, which require random projections in addition to random permutations.

There are many promising extensions. For example, we can construct new kernels based on CoRE kernels (which currently do not have tuning parameters), by using the exponential function and  introducing an additional tuning parameter $\gamma$, just like RBF kernel. This will allow more flexibility and potentially further improve the performance.

\clearpage

Another interesting  line of extensions would be  applying other hashing algorithms on our generated hashed data. This is possible again because we can view our estimators as inner products and hence we can apply other hashing algorithms which approximate inner products on top of our hashed data. The advantage is the potential further data compression. Another advantage would be in the context of sublinear time approximate near neighbor search (when the target similarity is the CoRE kernels).

For example, we can apply another layer of random projections on top of the hashed data and then store the signs of the new projected data~\cite{Proc:Charikar,Article:Goemans_JACM95}. These signs, which are bits, provide good indexing \& space partitioning capability to allow sublinear time approximate near neighbor search under the framework of LSH~\cite{Proc:Indyk_STOC98}. This way, we can search for near neighbors in the space of CoRE kernels (instead of inner products).

\section{Conclusion}

Current popular hashing methods, such as random projections and variants, often focus on approximating  inner products and large-scale linear classifiers (e.g., linear SVM). However, linear kernels often do not achieve  good performance. In this paper, we propose two types of CoRE kernels which outperform linear kernels, sometimes by a large margin, on sparse non-binary data (which are common in practice). Because CoRE kernels are nonlinear, we accordingly develop new hash methods to approximate CoRE kernels. The hashed data can be fed into highly efficient linear classifiers. Our experiments confirm the findings. We expect this work will inspire a new line of research on hashing algorithms and large-scale learning.

\appendix
\section{Proof of Theorem~\ref{thm_Kc1}}\label{app_thm_Kc1}

To compute the expectation and variance of the estimator  $\hat{K}_{C,1} = \frac{1}{k}\sum_{j=1}^k P_j(u)P_j(v)1\{L_j(u)=L_j(v)\}$, we  need  the first two moments
of $P_j(u)P_j(v)1\{L_j(u)=L_j(v)\}$. The first moment is
\begin{align}\notag
&E\left[P_j(u)P_j(v)1\{L_j(u)=L_j(v)\}\right]\\\notag
=& E\left[P_j(u)P_j(v)\right]\mathbf{Pr}\left(L_j(u)=L_j(v)\right) = \rho R
\end{align}
which implies that $E\left(\hat{K}_{C,1}\right) = K_{C,1} = \rho R$.  The second moment is
\begin{align}\notag
&E\left[P_j^2(u)P_j^2(v)1\{L_j(u)=L_j(v)\}\right]\\\notag
=&E\left[P_j^2(u)P_j^2(v)\right] \mathbf{Pr}\left(L_j(u)=L_j(v)\right)\\\notag
=& \left(1+2\rho^2\right) \rho R
\end{align}
Here, we have used the result in the prior work~\cite{Proc:Li_Hastie_Church_KDD06}: $E\left[P_j^2(u)P_j^2(v)\right] = 1+2\rho^2$. Therefore, the variance is
\begin{align}\notag
Var\left(\hat{K}_{C,1}\right) =\frac{1}{k}\left\{\left(1+2\rho^2\right)R -  \rho^2 R^2\right\}
\end{align}

This completes the proof.

\section{Proof of Theorem~\ref{thm_Kc2}}\label{app_thm_Kc2}
We need  the first two moments of the  estimator $\hat{K}_{C,2} = \frac{1}{k}\sum_{j=1}^k V_j(u) V_j(v)1\{L_j(u) = L_j(v)\}\sqrt{f_1f_2}$

Because
\begin{align}\notag
&E\left[V_j(u) V_j(v)1\{L_j(u) = L_j(v)\}\right]\\\notag
=&E\left[V_j(u) V_j(v)1\{L_j(u) = L_j(v)\}|L_j(u)=L_j(v)\right]\\\notag
&\times\mathbf{Pr}\left(L_j(u)=L_j(v)\right)\\\notag
=&\frac{\sum_{i=1}^D u_iv_i}{a}R
=\rho \frac{1}{f_1+f_2-a}
\end{align}
we know
\begin{align}\notag
E\left(\hat{K}_{C,2}\right) = \frac{1}{k}\sum_{j=1}^k \rho\frac{\sqrt{f_1f_2}}{f_1+f_2-a} = K_{C,2}
\end{align}
and
\begin{align}\notag
&E\left[V_j^2(u) V_j^2(v)1\{L_j(u) = L_j(v)\}\right]\\\notag
=&E\left[V_j^2(u) V_j^2(v)\right]\mathbf{Pr}\left(L_j(u)=L_j(v)\right)\\\notag
=&\frac{\sum_{i=1}^D u_i^2v_i^2}{a}R
=\frac{\sum_{i=1}^D u_i^2v_i^2}{f_1+f_2-a}
\end{align}
Therefore,
\begin{align}\notag
&Var\left(\hat{K}_{C,2}\right) \\\notag
=& \frac{1}{k}\frac{f_1f_2}{f_1+f_2-a}\left(\sum_{i=1}^D u_i^2v_i^2-\frac{\left(\sum_{i=1}^D u_iv_i\right)^2}{(f_1+f_2-a)}\right)
\end{align}

This completes the proof.


\end{document}